%% file: main.tex
\documentclass[10pt,twocolumn,letterpaper]{article}

\usepackage[pagenumbers]{cvpr} %
\input{preamble}

\definecolor{cvprblue}{rgb}{0.21,0.49,0.74}
\usepackage[pagebackref,breaklinks,colorlinks,allcolors=cvprblue]{hyperref}

\def\paperID{325} 
\def\confName{CVPR}
\def\confYear{2026}

\title{\ourname{}: Zero-shot 3D Completion with Latent–Spatial Consistency }

\author{
Weilong Yan$^{1}$ \qquad
Haipeng Li$^{2}$ \qquad
Hao Xu$^{3}$ \qquad
Nianjin Ye$^{4}$\\
Yihao Ai$^{1}$\qquad
Shuaicheng Liu$^{2}$\qquad
Jingyu Hu$^{3\dagger}$\\[4pt]
{\small
$^{1}$National University of Singapore \quad
$^{2}$University of Electronic Science and Technology of China}
\\
{\small
$^{3}$The Chinese University of Hong Kong \quad
$^{4}$Changhong Intelligent Robot
}
}

\begin{document}
\maketitle
\renewcommand\thefootnote{} 
\footnotetext{$\dagger$ Corresponding author}

\input{sec/0_abstract}    
\input{sec/1_intro}

\input{sec/2_rw}
\input{sec/3_method}
\input{sec/4_exp}

\input{sec/5_conclusion}
\clearpage
\onecolumn
\input{Supp/X_suppl}

\twocolumn
\clearpage
{
    \small
    \bibliographystyle{ieeenat_fullname}
    \bibliography{main}
}


\end{document}

%% file: preamble.tex
\newcommand{\ourname}[1]{LaS-Comp}

\def \eg {{\emph{e.g.}\thinspace}}

\usepackage{pifont}
\usepackage{booktabs}
\usepackage{threeparttable} 

\usepackage{tabularx,booktabs,makecell,array,multirow,xcolor}
\newcolumntype{Y}{>{\centering\arraybackslash}X}  

\usepackage{capt-of}
\usepackage{float}

\usepackage{tabularx,booktabs,makecell,array}
\usepackage[table]{xcolor}      
\newcolumntype{L}{>{\raggedright\arraybackslash}X}

\newcolumntype{C}{>{\centering\arraybackslash}X}
\usepackage[table]{xcolor}
\usepackage{colortbl} 
\definecolor{bestpink}{HTML}{f9dce0}      
\definecolor{secondbeige}{HTML}{fff0d8}   

\usepackage{xcolor}
\newcommand{\bestcap}[2]{%
  \begingroup\fboxsep=0.6pt%
  \ifstrempty{#1}{%
    \colorbox{bestpink}{\strut\hspace{1.2em}}
  }{%
    \colorbox{bestpink}{\strut\textbf{#1#2}}
  }%
  \endgroup%
}
\newcommand{\secondcap}[2]{%
  \begingroup\fboxsep=0.6pt%
  \ifstrempty{#1}{%
    \colorbox{secondbeige}{\strut\hspace{1.2em}}%
  }{%
    \colorbox{secondbeige}{\strut{#1#2}}%
  }%
  \endgroup%
}

\usepackage{soul}
\usepackage{color}

\newcommand{\hlbest}[1]{{\sethlcolor{bestpink}\textbf{\hl{#1}}}}
\newcommand{\hlsecond}[1]{{\sethlcolor{secondbeige}\hl{#1}}}

%% file: sec/0_abstract.tex
\begin{abstract}

This paper introduces \ourname{}, a zero-shot and category-agnostic approach that leverages the rich geometric priors of 3D foundation models to enable 3D shape completion across diverse types of partial observations.
Our contributions are threefold:
First, \ourname{} harnesses these powerful generative priors for completion through a complementary two-stage design:
(i) an explicit replacement stage that preserves the partial observation geometry to ensure faithful completion; and (ii) an implicit refinement stage ensures seamless boundaries between the observed and synthesized regions.
Second, our framework is training-free and compatible with different 3D foundation models.
Third, we introduce Omni-Comp, a comprehensive benchmark combining real-world and synthetic data with diverse and challenging partial patterns, enabling a more thorough and realistic evaluation. 
Both quantitative and qualitative experiments demonstrate that our approach outperforms previous state-of-the-art approaches.
Our code and data will be available at \href{https://github.com/DavidYan2001/LaS-Comp}{LaS-Comp}.
\end{abstract}

%% file: sec/1_intro.tex
\section{Introduction}
\label{sec:intro}

Shape completion is a fundamental problem in 3D vision and graphics, aiming to reconstruct complete 3D shapes from partial observations, with broad applications in robotics \cite{zero-grasp, octmae}, autonomous driving \cite{syn2real-depth, RS-HM, zhang2024dcpi}, and AR/VR \cite{wonderworld, cao2024ssnerf, bridgingdaynighttargetclass}.
An effective shape completion approach should meet the following requirements:
(i) robustly handle diverse partial patterns, from single-view scans to missing semantic parts;
(ii) well generalize across broad object categories; 
(iii) avoid dependence on paired partial–complete datasets; and
(iv) flexibly support both text guidance and automatic completion without user input for applications.

Traditional methods~\cite{thrun2005shape, pauly2005example, shen2012structure, sung2015data, pauly2008discovering} use hand-crafted priors to infer the missing geometry from partial observations.
However, such priors lack flexibility and fail to generalize to diverse real-world shapes.
Subsequently, \cite{yuan2018pcn, dai2017shape} propose to learn a mapping from partial to complete shapes by training a neural network.
Yet, these methods rely on paired partial-complete datasets, 
limiting their generalizability to unseen categories.
To overcome such constraints, recent works~\cite{sdscomplete, compc, genpc} leverage generative priors from large pre-trained models to enable category-agnostic shape completion. 
However, these methods assume that a partial input can be rendered into at least one complete image, where the object appears geometrically complete in the rendered view. 
When this assumption fails,~\eg, the incomplete regions are visible from any viewpoint (see \cref{fig:teaser}(c)), the incomplete renderings lead to degraded results.

\input{figures/fig_teaser}

Recent 3D foundation models~\cite{zhao2025hunyuan3d, zhang2024clay, li2025triposg, he2025triposf, xiang2025structured, wu2025direct3ds2gigascale3dgeneration}, trained on large-scale datasets~\cite{deitke2023objaverse, deitke2023objaversexl}, show strong cross-category and in-the-wild shape generation capabilities, enabling a wide range of downstream applications such as shape editing~\cite{hu2024_cnsedit, hu2023clipxplore, hui2022neural, hu2026pegasus3dpersonalizationgeometry, hui2022template, cao20253dot} and analysis~\cite{du2025hierarchical, 4dpcchat, chen2022single}.
In contrast, existing shape completion methods struggle to generalize across the diverse input partial patterns and shape categories. 
This gap motivates us to exploit the powerful geometric priors embedded in these foundation models for more robust and generalizable shape completion.
Yet, unlike earlier 3D generative models~\cite{luo2021diffusion, zhou20213d,hui2022neural,hu2023neural, liu2023exim} that operate directly in the spatial domain,~\eg, points or voxels, these foundation models adopt a latent-generative-based pipeline:
A variational autoencoder first maps shape to a compact latent space, and a diffusion or flow-matching model is trained over this space.
While this design effectively compresses 3D data and supports scalable training of large foundation models, it simultaneously introduces a unique challenge for shape completion.
A straightforward approach for shape completion within the latent framework is to encode the partial input into the latent space and directly use the latent codes of these observed regions as conditions to guide the generative model to predict the missing geometry. However, we observe that even when a complete shape and its partial input share identical geometry in the overlapping regions, their latent encodings in those regions differ significantly. Therefore, directly completing shapes in latent space becomes unreliable due to this domain gap.

To address this challenge, we propose a zero-shot \textbf{Latent–Spatial Consistency Completion (\ourname{})} framework,
which bridges the gap between the latent and spatial domains of pre-trained 3D foundation models for faithful and category-agnostic shape completion.
\ourname{} operates via a complementary two-stage design:
First, we introduce an \textbf{Explicit Replacement Stage (ERS)}, ensuring fidelity to the input partial shape by directly injecting its geometric information into the latent representation.
Second, we adopt an \textbf{Implicit Alignment Stage (IAS)} to improve smoothness between observed and synthesized regions by refining the latent features with a geometry-alignment loss.
By combining explicit geometric replacement with implicit latent refinement, \ourname{} unleashes the power of 3D foundation models to deliver high-fidelity shape completions.

Our framework is compatible with different latent-generative-based 3D foundation models~\cite{xiang2025structured, wu2025direct3ds2gigascale3dgeneration}. 
By exploiting the rich geometric priors encoded by these models, our framework is capable of robustly handling diverse partial patterns, ranging from single-view scans to irregularly missing regions; see \cref{fig:teaser}(a-c). 
Furthermore, our approach leverages the built-in classifier-free guidance (CFG) mechanism of these foundation models to support both text-guided and unconditional shape completion, offering flexible user control; see \cref{fig:teaser}(d).
Finally, we found the evaluation gap in existing benchmarks~\cite{redwood, compc}, which are mostly limited to single-view partial scans and do not reflect real-world diversity.
To bridge this gap, we introduce Omni-Comp: a new benchmark composed of real-world scans and synthetic data with diverse, challenging partial patterns, enabling a more comprehensive in-the-wild evaluation.

In summary, our contributions are listed as follows:
\begin{itemize}
    \item We introduce \ourname{}, a novel zero-shot, category-agnostic 3D shape completion framework. 
    \ourname{} bridges the gap between latent and spatial domains via a two-stage design: the Explicit Replacement Stage (ERS) preserves fidelity to the partial input, while the Implicit Alignment Stage (IAS) ensures boundary coherence.
    \item Our framework is training-free, compatible with different 3D foundation models, and highly efficient. 
    It completes each shape in 20 seconds, making it over 3$\times$ faster than existing zero-shot methods.
    \item We introduce Omni-Comp, a new benchmark composed of both real-world scans and synthetic shapes with diverse partial patterns to enable more comprehensive in-the-wild evaluation. Extensive experiments on this benchmark and others confirm our method's state-of-the-art performance.
\end{itemize}

%% file: figures/fig_teaser.tex
\begin{figure}[!t]
    \centering
    \includegraphics[width=1.\linewidth]{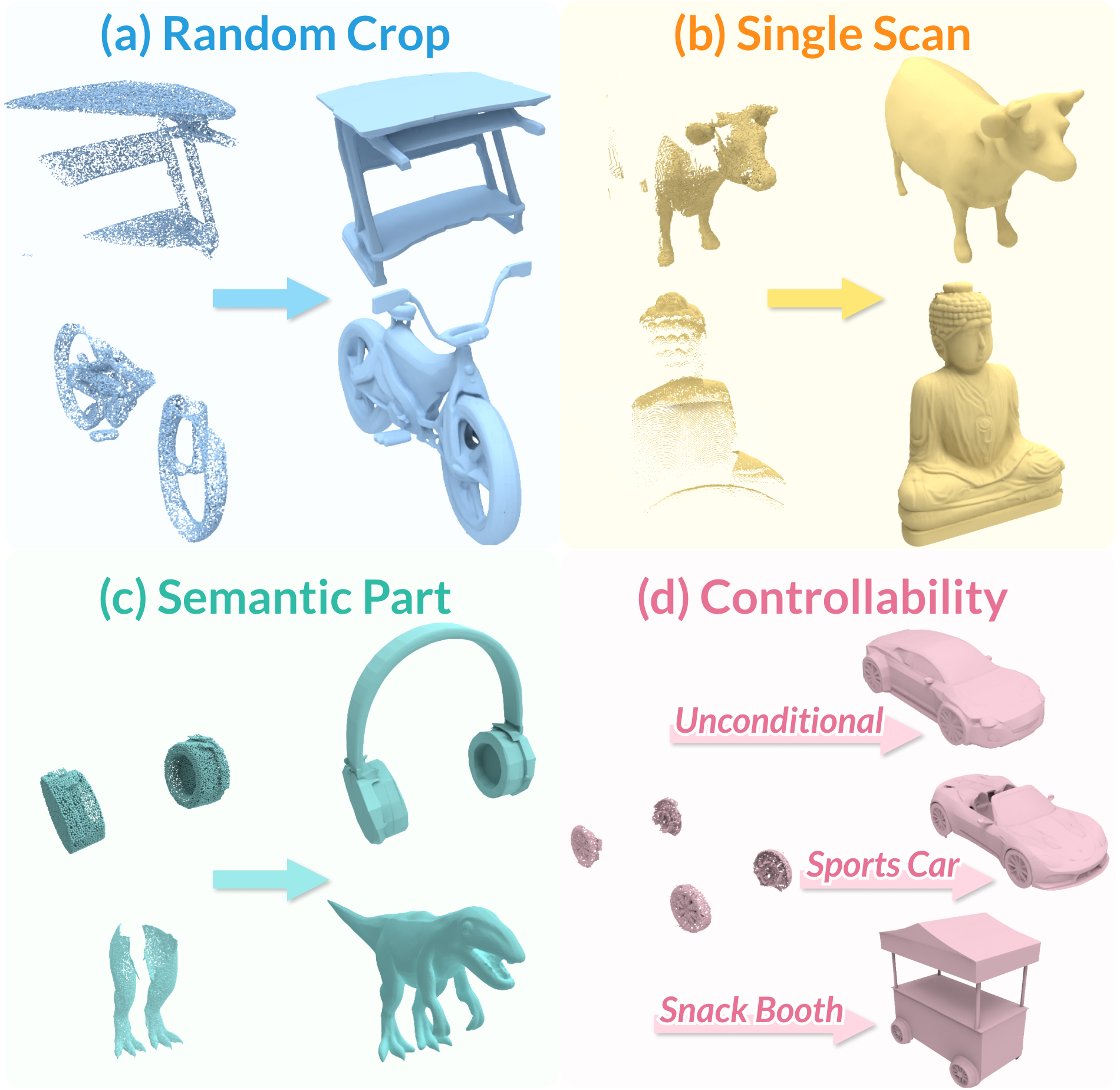}
    \vspace{-3mm}
    \caption{Our new framework supports category-agnostic shape completion across diverse partial patterns, including (a) random crops, (b) single-view scans, and (c) missing semantic parts. It further supports both unconditional and text-guided completion, offering flexible control for real-world applications, see (d).}
    \label{fig:teaser}
    \vspace{-3mm}
\end{figure}

%% file: sec/2_rw.tex
\section{Related Work}
\label{sec:rw}

\paragraph{Supervised Shape Completion.}
Supervised shape completion methods learn to reconstruct complete 3D shapes from partial inputs by training on paired datasets of partial and complete shapes.
Pioneering works~\cite{yuan2018pcn, dai2017shape} established this paradigm by introducing paired datasets and end-to-end networks to directly learn the mapping from partial to complete shapes.
Subsequent studies~\cite{zhang2020detail, xie2020grnet, huang2020pf, wang2020cascaded, wen2021pmp} employed coarse-to-fine refinement to better reconstruct missing regions and recover fine geometric details.
Transformer-based approaches~\cite{yu2021pointr, xiang2021snowflakenet, zhu2023svdformer, wang2024pointattn, li2023proxyformer, yan2022shapeformer, pcdreamer} leverage attention mechanisms to effectively aggregate multi-scale geometric features, leading to higher-fidelity shape completions.
Meanwhile, diffusion-based methods~\cite{chu2023diffcomplete, zhou20213d, chu2025digging, lyu2021conditional} formulate shape completion as a conditional generation task, where the partial input serves as a condition for the diffusion model~\cite{ho2020denoising} to produce a completed shape.
Although effective on in-domain data, these methods fail to generalize to unseen categories and partial patterns.
Moreover, collecting large-scale paired datasets is costly, motivating the exploration of unsupervised or self-supervised alternatives.

\input{figures/fig_pipeline}

\vspace{-4mm}
\paragraph{Unsupervised Shape Completion.}
To alleviate reliance on paired data, recent approaches explore unsupervised shape completion. \cite{chen2019unpaired} first leverages unpaired data and adversarial losses. Follow-up works \cite{wen2021cycle4completion, shapeinv, cai2022learning, wu2020multimodal, xie2021style} further advance research through various unsupervised strategies, e.g., cycle consistency and latent space alignment. Building upon these efforts, recent self-supervised approaches \cite{cui2023p2c, hong2023acl, chu2021unsupervised, liu2024self, kim2023learning} relax the need for complete shapes by learning geometric priors directly from partial observations. However, these methods are trained on ShapeNet \cite{chang2015shapenet}, which contains limited categories, leading to degraded performance on real-world data or unseen categories.

\paragraph{Shape Completion with Generative Priors.} 
To overcome the limitations of paired-data supervision, recent works~\cite{sdscomplete, compc, genpc} leverage priors from large-scale pre-trained generative models for category-agnostic 3D shape completion.
SDS-Complete~\cite{sdscomplete} distills 2D priors from Stable Diffusion~\cite{rombach2022high} to optimize the signed distance field, while ComPC~\cite{compc} extends this idea by initializing partial inputs as 3D Gaussians and utilizing Zero-1-to-3~\cite{liu2023zero} to render multi-view images and iteratively refine them for shape completion.
GenPC~\cite{genpc} renders partial input into an image, reconstructs a coarse shape via image-to-3D models~\cite{hong2024lrmlargereconstructionmodel, tang2024lgm}, and refines it using Zero-1-to-3~\cite{liu2023zero}.
However, these methods rely on the assumption that partial inputs can be rendered into at least one reasonably complete image. 
When the partial input fails to yield a complete image from any viewpoint, the incomplete renderings often cause suboptimal completion results.
To address the limitations of existing approaches, we propose the first framework that fully leverages latent-generative-based 3D foundation models for high-fidelity, zero-shot, and category-agnostic shape completion. Without relying on any rendering assumptions, our method robustly handles highly diverse partial input patterns, from single-view scans to irregular occlusions. Moreover, it is fully training-free, enabling seamless deployment across a wide range of real-world scenarios.

%% file: figures/fig_pipeline.tex
\begin{figure*}[t]
    \centering
    \includegraphics[width=1\linewidth]
    {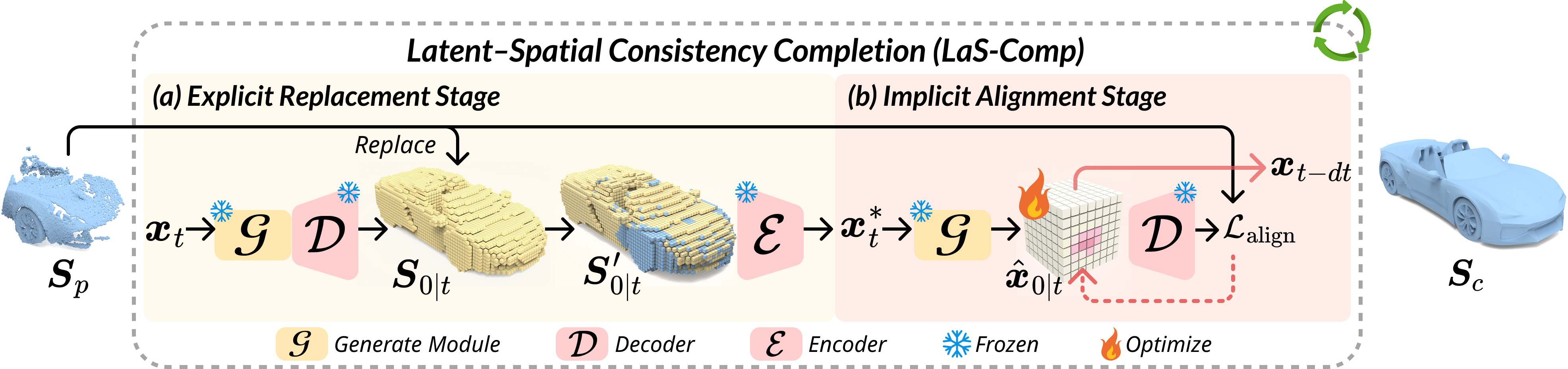}
    \vspace{-5mm}
    \caption{Overview of the \ourname{} framework. Starting from Gaussian noise, the process iteratively refines a latent feature $\boldsymbol{x}_t$ under the guidance of the partial input $\boldsymbol{S}_p$. At each iteration $t$, this refinement is performed in two stages: the \textit{Explicit Replacement Stage (ERS)} and the \textit{Implicit Alignment Stage (IAS)}. The ERS explicitly injects the known geometry of $\boldsymbol{S}_p$ into $\boldsymbol{x}_t$ to produce an updated latent $\boldsymbol{x}^{*}_t$. The IAS then refines $\boldsymbol{x}^{*}_t$ using a gradient-based optimization, yielding a spatially aligned latent $\boldsymbol{x}_{t-dt}$ for the next step. After the final iteration, the completed shape $\boldsymbol{S_{\text{c}}}$ is obtained by decoding the refined latent.
    \label{fig:pipeline}} 
    \vspace{-0.3cm}
\end{figure*}

%% file: sec/3_method.tex
\section{Method}

\subsection{Overview}
Given a partial 3D shape $\boldsymbol{S}_p\in\mathbb{R}^{k \times 3}$, we desire to generate a completed shape $\boldsymbol{S}_{c}\in\mathbb{R}^{k \times 3}$ that is geometrically faithful to $\boldsymbol{S}_p$.
To achieve this, we introduce the \ourname{} framework, which effectively leverages the powerful geometric priors of pre-trained 3D foundation models~\cite{xiang2025structured, wu2025direct3ds2gigascale3dgeneration} for zero-shot shape completion. 
Through the classifier-free guidance sampling mechanism~\cite{ho2022classifierfreediffusionguidance} of 3D foundation models, our method supports both text-guided and unconditional shape completion.

As illustrated in \cref{fig:pipeline}, starting from a Gaussian noise, our framework iteratively refines it through multiple denoising steps under the guidance of the partial input $\boldsymbol{S}_p$, progressively recovering the underlying complete geometry.
At each iteration $t\!\in\![0, 1]$, \ourname{} takes the current latent feature $\boldsymbol{x}_t\in\mathbb{R}^{n^3 \times c}$ as input, together with the partial shape $\boldsymbol{S}_p$, and performs two complementary operations: 
\begin{itemize}
    \item \textit{Explicit Replacement Stage (ERS)} explicitly injects the partial input $\boldsymbol{S}_p$ into $\boldsymbol{x}_t$, producing an updated latent $\boldsymbol{x}^{*}_t$ that enforces strict fidelity to the partial input.
    \item \textit{Implicit Alignment Stage (IAS)} refines $\boldsymbol{x}^{*}_t$ through a one-step optimization guided by a geometric-alignment loss, which encourages the smoothness between synthesized and observed regions, producing a coherent latent $\boldsymbol{x}_{t-dt}$ for the next step ($dt$ is the length of a denoising step).
\end{itemize}

After the final iteration, the refined latent feature $\boldsymbol{x}_0$ is decoded through the decoder $\mathcal{D}$ to produce the completed shape: $\boldsymbol{S}_c = \mathcal{D}(\boldsymbol{x}_0)$.
The details of ERS and IAS are demonstrated in \cref{sec:ers} and \cref{sec:ias}, respectively.

\subsection{Explicit Replacement Stage}
\label{sec:ers}
Taking the latent feature $\boldsymbol{x}_t$ from the previous generative step and the partial shape $\boldsymbol{S}_p$ as input, the ERS performs primary completion by injecting the geometry of $\boldsymbol{S}_p$ into the completion process, producing an updated latent feature $\boldsymbol{x}^{*}_t$ that explicitly encodes the observed regions.

As depicted in \cref{fig:ERS}, motivated by the latent decomposition in FlowDPS~\cite{Kim_2025_ICCV}, we decompose the generation at each timestep into two branches to enhance both fidelity and diversity: (i) a clean branch for enforcing the input fidelity; and (ii) a noisy branch for improving generation diversity.

\vspace{-3mm}
\paragraph{Clean Branch.}
Given the current latent feature $\boldsymbol{x}_t$, the clean branch uses the generator $\mathcal{G}$ to estimate the noise-free latent feature $\hat{\boldsymbol{x}}_{0|t}$ based on the linear flow path:
\begin{equation}
    \hat{\boldsymbol{x}}_{0|t} = \boldsymbol{x}_t - t \cdot \mathcal{G}(\boldsymbol{x}_t, t),
    \label{eq:clean_branch}
\end{equation}
where $\mathcal{G}(\boldsymbol{x}_t, t)$ is the predicted velocity from $\boldsymbol{x}_t$ towards $\boldsymbol{x}_0$.
This latent feature $\hat{\boldsymbol{x}}_{0|t}$ is then decoded into the spatial domain to yield a complete shape prediction:
\begin{equation}
    \boldsymbol{S}_{0|t} = \mathcal{D}(\hat{\boldsymbol{x}}_{0|t}).
\end{equation}

\input{figures/fig_ers}

Although $\boldsymbol{S}_{0|t}\in \mathbb{R}^{N^3}$ provides a complete shape produced by $\mathcal{G}$, it does not yet faithfully preserve the observed regions of the partial input $\boldsymbol{S}_p$.
To enforce this fidelity, we first voxelize the input partial shape $\boldsymbol{S}_p$  into an occupancy grid of size $N^3$ aligned with the spatial domain of $\boldsymbol{S}_{0|t}$. We then construct a binary spatial mask 
$\boldsymbol{M} \!\in\! \{0, 1\}^{N^3}$
, where $\boldsymbol{M}\!=\!1$ indicates occupied voxels observed in $\boldsymbol{S}_p$. 
Using this mask, we perform a spatial replacement to inject the known geometry into the generative prediction:
\begin{equation}
    \boldsymbol{S}^{\prime}_{0|t} = \boldsymbol{S}_p \odot \boldsymbol{M} + \boldsymbol{S}_{0|t} \odot (1 \!-\! \boldsymbol{M}),
\end{equation}
where $\odot$ is element-wise multiplication.
This operation explicitly replaces the geometry with $\boldsymbol{S}_p$ for fidelity.
The missing regions are then completed by the generative prediction $\boldsymbol{S}_{0|t}$.
Next, we feed $\boldsymbol{S}^{\prime}_{0|t}$ into the encoder $\mathcal{E}$ to produce a latent feature $\boldsymbol{x}^{*}_{0|t}$ that incorporates the geometries from the partial input as follows:
\begin{equation}
    \boldsymbol{x}^{*}_{0|t} = \mathcal{E}(\boldsymbol{S}^{\prime}_{0|t}).
    \label{eq:noise_branch}
\end{equation}

\vspace{-3mm}
\paragraph{Noisy Branch.}
Concurrently, the noisy branch estimates the noisy latent $\hat{\boldsymbol{x}}_{1|t}$ using the same generator prediction $\mathcal{G}(\boldsymbol{x}_t, t)$ as in \cref{eq:clean_branch}:
\begin{equation}
    \hat{\boldsymbol{x}}_{1|t} = \boldsymbol{x}_t + (1\!-\!t) \cdot \mathcal{G}(\boldsymbol{x}_t, t).
\end{equation}

Rather than employing $\hat{\boldsymbol{x}}_{1|t}$ for latent update, we introduce a novel \textit{Partial-aware Noise Schedule (PNS)} to modulate the stochasticity during denoising.
The motivation is that $\hat{\boldsymbol{x}}_{1|t}$ treats all spatial regions equally, whereas shape completion is asymmetric: observed regions should remain stable to preserve the input geometry, while missing regions should permit higher stochasticity to explore plausible completions.
The final noisy latent $\boldsymbol{x}^{*}_{1|t}$ is then composed as:
\begin{equation}
\label{PAS}
    \boldsymbol{x}_{_{1|t}}^*=\boldsymbol{M} \odot \left( \sqrt{1\!-\!t} \cdot \hat{\boldsymbol{x}}_{1|t} + \sqrt{t} \cdot \boldsymbol{\epsilon}_1 \right) + (1\!-\!\boldsymbol{M}) \odot \boldsymbol{\epsilon}_2,
\end{equation}
where $\boldsymbol{\epsilon}_1, \boldsymbol{\epsilon}_2 \sim \mathcal{N}(\boldsymbol{0}, \boldsymbol{I})$ and $\boldsymbol{M}$ is downsampled to size of $n^3$ to match the dimensions of the latent features. 
This design applies the following mechanisms:
\begin{itemize}
    \item For the observed regions ($\boldsymbol{M}\!=\!1$): 
    Since these regions correspond to the reliable partial input, their latent features should remain largely stable during denoising, with only minimal stochastic disturbance.
    To achieve this, we blend the model-predicted noisy latent $\hat{\boldsymbol{x}}_{1|t}$ with Gaussian noise $\boldsymbol{\epsilon}_1$ using the coefficients $\sqrt{1-t}$ and $\sqrt{t}$.
    This schedule imposes a time-dependent perturbation magnitude: early iterations inject greater randomness to allow adjustments that maintain overall coherence, while later iterations gradually reduce stochasticity and preserve the geometry of the observed regions.
    \item For the missing regions ($\boldsymbol{M}\!=\!0$):
    No reliable observations are available to constrain the denoising process.
    To encourage broad exploration of the possible completions, we replace the model-predicted noisy latent with pure Gaussian noise $\boldsymbol{\epsilon}_2$.
    This promotes diversity in the generated geometry for the missing areas.
\end{itemize}

With the updated noise-free latent $\boldsymbol{x}^{*}_{0|t}$ and the noisy latent $\boldsymbol{x}^{*}_{1|t}$, we reconstruct the updated latent state $\boldsymbol{x}^{*}_t$ for the current timestep $t$ via the forward flow interpolation:
\begin{equation}
    \boldsymbol{x}^{*}_t = (1\!-\!t) \cdot \boldsymbol{x}^{*}_{0|t} + t \cdot \boldsymbol{x}^{*}_{1|t},
    \label{eq:reconstruct_xt}
\end{equation}
the resulting latent  $\boldsymbol{x}^{*}_t$ encodes the partial input geometry via the ERS, where the observed regions are directly imposed in the spatial domain. Although this explicit spatial replacement preserves fidelity to the input, it may introduce inconsistencies or discontinuities near the boundaries between observed and synthesized regions. To mitigate these artifacts and improve local coherence, we feed $\boldsymbol{x}^{*}_t$ into the Implicit Alignment Stage (IAS) for a refinement step.

\subsection{Implicit Alignment Stage}
\label{sec:ias}
To improve coherence between observed and synthesized regions, we introduce the Implicit Alignment Stage (IAS), which refines the ERS output $\boldsymbol{x}^{*}_t$ before the next iteration.
The refinement begins by estimating a noise-free latent from $\boldsymbol{x}^{*}_t$.
Refer to the \cref{eq:clean_branch}, we reuse the notation $\hat{\boldsymbol{x}}_{0|t}$ to denote the noise-free latent predicted from $\boldsymbol{x}^{*}_t$:
\begin{equation}
    \hat{\boldsymbol{x}}_{0|t} = \boldsymbol{x}^{*}_t - t \cdot \mathcal{G}(\boldsymbol{x}^{*}_t, t),
    \label{eq:ias_clean_decomp}
\end{equation}
then this predicted noise-free latent $\hat{\boldsymbol{x}}_{0|t}$ is then decoded into its spatial domain to obtain $\boldsymbol{S}_{0|t}$:
\begin{equation}
    \boldsymbol{S}_{0|t} = \mathcal{D}(\hat{\boldsymbol{x}}_{0|t}).
\end{equation}

To reduce the boundary artifacts introduced by the ERS, we define geometry-alignment loss $\mathcal{L}_{\text{align}}$ to refine the latent feature.
This loss provides a localized correction within the reliable masked regions, helping smooth the discontinuities.
Specifically, we use the mask $\boldsymbol{M}$ to select the masked regions and compute a BCE loss between the predicted occupancy and the corresponding voxels of the partial input:
\begin{equation}
    \mathcal{L}_{\text{align}} = \text{BCE}(\boldsymbol{S}_{0|t} \odot \boldsymbol{M}, \boldsymbol{S}_{p} \odot \boldsymbol{M}).
    \label{eq:ias_loss}
\end{equation}

Notice that this loss is not used to update the model parameters. 
Instead, we compute the gradient of the loss with respect to the latent feature, and optimize it via a \textbf{single-step} gradient update to refine it: 
\begin{equation}
    \boldsymbol{x}^{\text{aligned}}_{0|t} = \hat{\boldsymbol{x}}_{0|t} - \eta \cdot \nabla_{\hat{\boldsymbol{x}}_{0|t}} \mathcal{L}_{\text{align}},
    \label{eq:ias_optim}
\end{equation}
where $\eta$ is the learning rate.
By refining the feature $\hat{\boldsymbol{x}}_{0|t}$ to better match the reliable masked regions, we obtain an updated latent feature $\boldsymbol{x}^{\text{aligned}}_{0|t}$ that is more coherent.
This updated feature is then used to compute the latent state $\boldsymbol{x}_{t-dt}$ for the next iteration as follows:
\begin{equation}
    \boldsymbol{x}_{t-dt} = \boldsymbol{x}^{\text{aligned}}_{0|t} + (t \!-\! dt) \cdot \mathcal{G}(\boldsymbol{x}^{*}_t, t).
    \label{eq:final_update}
\end{equation}

The latent feature $\boldsymbol{x}_{t-dt}$ then serves as the input for the next iteration of the 
completion process.

%% file: figures/fig_ers.tex
\begin{figure}[t]
    \centering
    \includegraphics[width=1\linewidth]
    {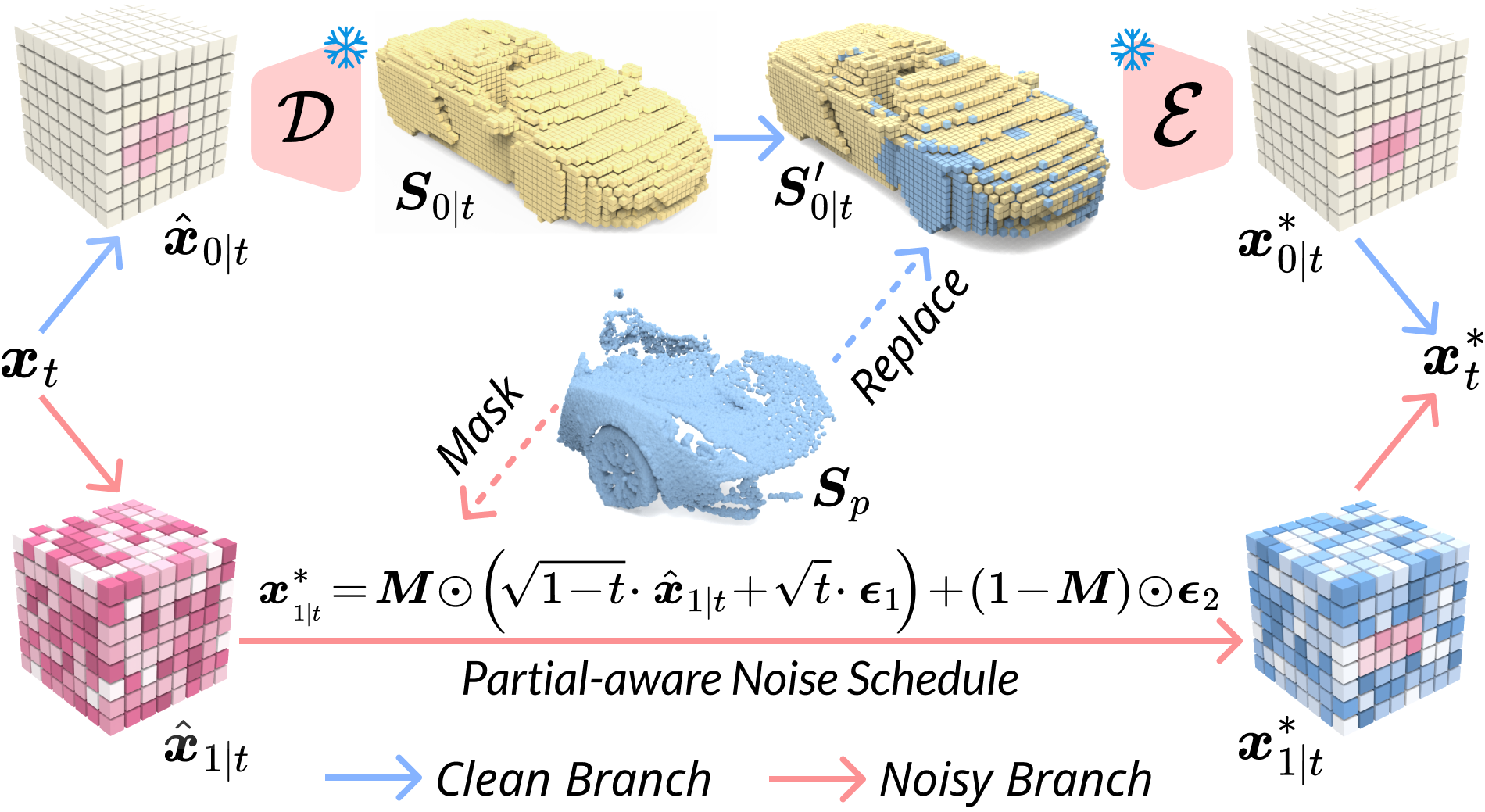}
    \vspace{-0.5cm}
    \caption{Overview of the {Explicit Replacement Stage (ERS)}. At each timestep $t$, ERS decomposes the latent generation into two parallel branches. The {clean branch} (top) enforces spatial consistency, yielding $\boldsymbol{x}^{*}_{0|t}$. Concurrently, the {noisy branch} (bottom) enhances fidelity, producing $\boldsymbol{x}^{*}_{1|t}$. These two branch outputs are then interpolated to compute the final aligned latent $\boldsymbol{x}^{*}_t$.}
    \label{fig:ERS}
    \vspace{-0.3cm}
\end{figure}

%% file: sec/4_exp.tex
\section{Experiments}

\input{Tables/table_redwood}

\input{figures/fig_compare_redwood} 

\subsection{Datasets and Implementation details}
\label{datasets and metrics}

\paragraph{Evaluation Datasets.}
We conduct a comprehensive evaluation across multiple benchmarks.
(i) \textbf{Redwood}~\cite{redwood}: We follow \cite{sdscomplete, compc, genpc}, utilizing 10 reconstructed meshes from real-world RGB-D scans, with single scans as the partial input.
(ii) \textbf{Synthetic dataset}~\cite{compc}: Following \cite{compc, lipman2008green}, this contains 12 objects from different categories, with virtually-rendered single scans as the partial input.
(iii) \textbf{KITTI}~\cite{dataset_kitti}: We follow \cite{chen2019unpaired, shapeinv}, using cars from real-world LiDAR scans, with extremely sparse points as the partial input.
(iv) \textbf{ScanNet}~\cite{dataset_scannet}: We follow \cite{chen2019unpaired, shapeinv, cui2023p2c}, cropping 48 chairs and 49 tables from real-world single scans (no ground truths).

\vspace{-5mm}
\paragraph{Proposed Omni-Comp Benchmark.}
These aforementioned benchmarks, while valuable, highlight clear limitations for comprehensive evaluation, including limited scales (\eg, about 10 meshes in Redwood/Synthetic), restricted category diversity (\eg, $\leq$2 categories in KITTI/ScanNet), and confinement to a single partial pattern (\eg, depth/LiDAR scans).
To address these limitations, we introduce \textbf{Omni-Comp}, a new benchmark designed for a more comprehensive and robust evaluation of 3D shape completion.
Our benchmark features a challenging set of 30 objects, each from a distinct category, curated from diverse sources: 10 real-world scans from Redwood~\cite{redwood} (chosen for complex geometry), 10 real-world everyday objects from YCB~\cite{ycb} (motivated by downstream applications, \eg, robotic grasping), and 10 synthetic shapes from \cite{partnextnext} (chosen for rich semantic structure).
Critically, inspired by \cite{shapeinv}, our benchmark generates three distinct partiality patterns for each object:
(i) \textit{Single Scan}: using the projection of the captured depth map for real-world data, and simulating a standard depth camera capture for synthetic data; 
(ii) \textit{Random Crop}: Representing arbitrary occlusions by randomly cropping a portion; and 
(iii) \textit{Semantic Part}: Keeping a semantic component and removing other parts. 
By creating two samples for each pattern per object, the benchmark comprises 180 partial samples with ground truths. 
More details are in the supplementary material.

\vspace{-5mm}
\paragraph{Evaluation Metrics.}
We follow previous works \cite{compc, shapeinv, sdscomplete, genpc} to assess completion quality from three perspectives: 
(i) Chamfer Distance (CD) and Earth Mover's Distance (EMD), reported both $\times\! 10^{2}$, which measures completion accuracy by the distance between the prediction and the ground truth;
%
(ii) Unidirectional Chamfer Distance (UCD) and Unidirectional Hausdorff Distance (UHD), reported $\times\! 10^{4}$ and $\times\! 10^{2}$, which measure the fidelity of the prediction compared with the partial input; and 
(iii) Minimum Matching Distance (MMD) and Total Mutual Difference (TMD), reported both $\times\! 10^{2}$, which assess completion diversity across multiple outputs. 
More details are in the supplementary material.

\vspace{-5mm}
\paragraph{Baselines.} 
We compare our method against a comprehensive set of recent approaches, categorized into three main groups, including supervised methods~\cite{pcdreamer, adapointr, zhu2023svdformer}, unsupervised methods~\cite{shapeinv, cui2023p2c}, and generative prior-based (zero-shot) methods~\cite{compc, genpc, sdscomplete}.
To ensure a fair comparison, we adopt the same experiment settings following previous zero-shot 3D completion methods~\cite{compc, genpc, sdscomplete, shapeinv}. 

\vspace{-5mm}
\paragraph{Implementation Details.}
We implement our framework based on two pre-trained 3D generative foundation models, TRELLIS~\cite{xiang2025structured} and Direct3D-S2~\cite{wu2025direct3ds2gigascale3dgeneration}.
%
Following \cite{compc}, we sample point clouds from the output meshes with farthest point sampling for quantitative evaluation. 
%
%
%
The learning rate for the one-step  IAS refinement step is set to $1\times10^{-5}$.
More details are in the supplementary material.

\subsection{Comparison with State-of-the-art Methods}
\label{quantitative exp}
We present comprehensive analyses across multiple benchmarks for assessment in this section.
For fair visualization, all point clouds are converted to meshes using~\cite{Peng2021SAP}.

\input{Tables/table_plyobj}

\vspace{-5mm}
\paragraph{Evaluation on Completion Correctness.}
We first evaluate the completion correctness on single-scan partial shapes from the real-world Redwood dataset~\cite{redwood} and synthetic data~\cite{compc, lipman2008green}, as shown in \cref{tab:exp_redwood_cd_emd} and \cref{tab:exp_synthetic_cd_emd}. 
Our method achieves state-of-the-art performance in most categories and demonstrates significant improvements compared to recent zero-shot methods. 
Specifically, we outperform ComPC~\cite{compc} by 27.2\% in CD and 29.0\% in EMD, and surpass GenPC~\cite{genpc} by 18.4\% in CD and 36.1\% in EMD, respectively.
We attribute this significant gap to a fundamental difference in methodology.
Prior methods~\cite{compc, genpc} either rely on generating multiple 2D views for supervision or lift a 2D rendered image to 3D with post-processing. 
In contrast, our approach directly conducts completion by leveraging both the strong generative priors and the rich 3D information within the partial input, enabling a more robust and accurate inference of the missing geometry.

We show qualitative comparisons in \cref{fig:viz_comp_redwood}.
For specific examples, such as the leaf structures of the plant (see Row (a)) and the rim and wheels of the trash bin (see Rows (b-c)), our method achieves high fidelity in the observed regions while generating missing parts with precise surface geometry and coherent topology.
In contrast, existing methods either distort the visible geometry or generate over-smoothed, implausible structures. 
This demonstrates that enforcing latent-spatial consistency with strong 3D generative priors yields superior geometric realism and structural integrity.

\input{Tables/table_kitti_scannet}

\vspace{-4mm}
\paragraph{Evaluation on Completion Fidelity.}
To test generalization and robustness to more challenging inputs, we evaluate on datasets with complex partiality, including sparse LiDAR scans from KITTI~\cite{dataset_kitti} and noisy depth scans from ScanNet~\cite{dataset_scannet}
As shown in \cref{tab:mmd_tmd_redwood_synthetic}, our method outperforms the latest zero-shot method ComPC~\cite{compc} by a large margin in almost all categories, confirming its robustness and superior fidelity on challenging real-world data.  Some qualitative examples can be found in \cref{fig:scannet_and_kitti}.

\input{Tables/table_omni_comp}

\input{figures/fig_scannet_kitti}
\input{figures/fig_compare_omni}  
\vspace{-4mm}
\paragraph{Evaluation on Omni-Comp.}
Our proposed Omni-Comp dataset specifically tests generalization across diverse partial patterns in both real-world and synthetic scenarios.
As shown in \cref{tab:cd_emd_three_patterns}, we report the average metrics on all shapes according to different partial patterns.
Previous unsupervised approaches~\cite{compc, sdscomplete} are heavily optimized for single-scan patterns, suffer a consistent and significant performance drop when generalizing to other partiality types. 
In contrast, our method leverages a pattern-agnostic 3D prior, enabling it to maintain robust and superior performance across all categories and patterns.
On average, we achieve improvements of 49.6\% in CD and 39.4\% in EMD over ComPC~\cite{compc}, demonstrating the effectiveness of our idea.

We provide some qualitative comparisons in \cref{fig:viz_comp_omni}.
Specifically, baselines' reliance on rendered 2D views that contain the complete object contour ~\cite{pcdreamer, compc} fails on complex inputs where contours are ill-posed (see Rows (c-d)). 
%
Furthermore,  \cite{pcdreamer, compc} fail to complete the random crop and semantic part due to out-of-distribution categories and partial patterns (see Rows (a-b) and (e-f)).
%
In stark contrast, our method proves robust across all these challenging cases, leveraging its pattern-agnostic 3D prior to reconstruct structurally plausible and geometrically detailed completions, demonstrating its strong generalizability.

\vspace{-5mm}
\paragraph{Evaluation on Completion Diversity.}
Finally, we assess completion diversity against the latest generative-based baselines~\cite{pcdreamer, compc} on Redwood~\cite{redwood} and synthetic~\cite{compc, lipman2008green} datasets, reporting average metrics over 5 random runs per object for each method. 
%
As shown in \cref{tab:mmd_tmd_redwood_synthetic}, our approach achieves consistently better MMD and TMD. 
Notably, while PCDreamer~\cite{pcdreamer} utilizes generative models, its reliance on varying multi-view depth maps produces limited geometric variation, resulting in low TMD scores. 
Our strong performance on both metrics verifies that our method can generate diverse shape completions, as shown in \cref{fig:diversity}.

\input{Tables/table_mmd_tmd}

\input{Tables/table_ablation}

\subsection{Ablation Studies}
\label{sec:ablation}
 We conduct ablation studies to validate the contribution of each component in our framework, and the quantitative and qualitative results are presented in \cref{tab:ablation_cd_emd} and \cref{fig:ablation}.

\vspace{-3mm}
\paragraph{Naive Baseline.}
We first establish a naive baseline that only utilizes the latents from the partial shape for replacement. 
Due to the latent discrepancy of the corresponding regions of partial and ground truth samples, it produces poor completion results both quantitatively and qualitatively, as shown in \cref{tab:ablation_cd_emd} (comparing Rows (a) and (f)) and \cref{fig:ablation}.

\vspace{-3mm}
\paragraph{Analysis of Key Components.}
We analyze the contribution of our three key components, Explicit Replacement Stage (ERS), Partial-aware Noise Schedule (PNS), and Implicit Alignment Stage (IAS), by ablating them individually.
\begin{itemize}
    \item \textbf{w/o ERS:} Comparing Rows (b) and (f) in \cref{tab:ablation_cd_emd}, removing the ERS causes the most significant performance drop. 
    Without this explicit geometric conditioning, the model fails to preserve the input structure and instead hallucinates a novel shape based on the generative prior (\eg, a completely different horse pose in \cref{fig:ablation}(c)). 
    This confirms that ERS is critical for ensuring input fidelity.
    
    \item \textbf{w/o PNS:} Comparing Rows (c) and (f) in \cref{tab:ablation_cd_emd}, disabling the PNS leads to noticeable stripe-like artifacts on the surface. 
    We attribute this to the mismatch in the denoising process between the clean (known) partial input and the noisy (unknown) regions being generated. 
    PNS is thus essential for seamlessly blending these two regions.
    
    \item \textbf{w/o IAS:} Comparing Rows (d) and (f) in \cref{tab:ablation_cd_emd}, removing the IAS, which harmonizes the latent context via optimization, results in small holes and boundary inconsistencies, as highlighted by red boxes in \cref{fig:ablation}.
    This demonstrates that ERS alone is insufficient; IAS is necessary to ensure the final latent representation is coherent.
    
    \item \textbf{Optimization steps in IAS:} We also do ablation about the optimization steps in IAS. As can be seen in \cref{tab:ablation_cd_emd} (e), using 10 steps for optimization at each timestep does not show obvious performance gain compared with (f).
\end{itemize}
\input{figures/fig_diversity}
\input{figures/fig_ablation}

The full pipeline successfully combines these components to achieve geometrically accurate and fidelity-preserving completions, with the best performance in \cref{tab:ablation_cd_emd}. The horse, as shown in \cref{fig:ablation} (f), exhibits excellent geometric quality without noticeable artifacts.

%% file: Tables/table_redwood.tex
\begin{table*}[!t]
\centering
\caption{
Quantitative comparisons on Redwood~\cite{redwood}.
We highlight the \bestcap{best}{} and \secondcap{second-best}{} results.
}
\vspace{-3mm}
\label{tab:exp_redwood_cd_emd}
\resizebox{\linewidth}{!}{
    \begin{tabular}{l|cccccccccc|c}
    \toprule
    \textbf{CD $\downarrow$ / EMD $\downarrow$} & \textit{Table} & \textit{Exe-Chair} & \textit{Out-Chair} & \textit{Old-Chair} & \textit{Vase} & \textit{Off-Can} & \textit{Vespa} & \textit{Tricycle} & \textit{Trash} & \textit{Couch} & \textit{Average} \\
    \midrule
    SVDFormer~\cite{zhu2023svdformer} &
    5.48/6.68 & 3.20/5.85 & \hlbest{0.79}/1.45 & 3.79/5.95 &
    5.70/6.98 & 5.13/6.71 & 3.10/4.91 & 2.73/4.96 &
    3.67/5.05 & 1.86/2.92 & 3.54/5.15 \\
    
    AdaPoinTr~\cite{adapointr} &
    5.02/6.25 & 2.58/4.80 & \hlsecond{0.82}/\hlbest{1.37} & 3.62/5.64 &
    5.14/6.50 & 4.47/6.35 & 1.96/3.54 & \hlsecond{1.83}/3.66 &
    \hlbest{1.21}/3.08 & \hlsecond{1.01}/\hlsecond{2.09} & 2.77/4.33 \\
    
    Shape-Inv~\cite{shapeinv} &
    1.58/2.84 & 3.59/5.75 & 1.36/2.12 & 4.55/7.39 &
    3.91/7.40 & 3.10/4.77 & 4.36/7.24 & 5.05/7.40 &
    2.48/3.83 & 1.95/2.76 & 3.19/5.15 \\
    
    P2C~\cite{cui2023p2c} &
    1.57/2.64 & 3.87/6.42 & 1.28/2.10 & 3.72/5.82 &
    4.54/7.04 & 3.36/4.82 & 6.75/10.7 & 7.78/11.2 &
    3.28/4.21 & 2.55/3.16 & 3.87/5.81 \\
    
    SDS-Comp~\cite{sdscomplete} &
    1.35/2.30 & 1.96/2.65 & 2.51/3.92 & 2.77/3.77 &
    3.00/5.25 & 3.79/4.28 & 3.36/5.73 & 3.18/3.49 &
    2.69/3.21 & 2.95/4.56 & 2.74/3.93 \\
    
    PCDreamer~\cite{pcdreamer} &
    \hlsecond{0.82}/2.36 & 1.43/3.56 & 1.19/2.07 & 2.61/5.34 &
    \hlsecond{2.41}/\hlsecond{3.64} & \hlsecond{2.62}/3.89 & 2.20/4.30 & 2.53/3.82 &
    1.55/2.47 & 1.47/2.61 & 1.88/3.41 \\
    
    GenPC~\cite{genpc} &
    1.28/2.07 & 1.43/2.29 & 1.16/1.68 & 1.36/2.20 &
    2.86/4.85 & 2.72/4.36 & 1.36/2.47 & \hlbest{1.38}/2.97 &
    2.31/3.17 & 1.58/2.78 & 1.74/2.88 \\
    
    ComPC~\cite{compc} &
    1.73/3.29 & \hlsecond{1.29}/\hlsecond{1.85} & 1.35/1.94 & 1.14/\hlsecond{1.63} &
    2.89/4.13 & 3.55/3.92 & \hlsecond{1.21}/\hlsecond{1.86} & 2.65/2.48 &
    2.15/2.34 & 1.59/2.58 & 1.95/2.59 \\
    
    \midrule
    
    Ours (Direct3D-S2) &
    \hlbest{0.75}/\hlbest{1.29} & 1.54/2.16 & 0.96/\hlbest{1.37} & \hlsecond{1.11}/1.69 &
    2.61/3.89 & 2.90/\hlsecond{2.92} & 1.35/2.08 & 1.96/\hlsecond{2.41} &
    \hlbest{1.21}/\hlsecond{1.85} & 1.95/2.26 & \hlsecond{1.64}/\hlsecond{2.19} \\
    
    Ours (TRELLIS) &
    \hlsecond{0.82}/\hlsecond{1.45} & \hlbest{1.25}/\hlbest{1.65} & 0.96/\hlsecond{1.38} & \hlbest{1.07}/\hlbest{1.37} &
    \hlbest{2.09}/\hlbest{3.61} & \hlbest{2.57}/\hlbest{2.53} & \hlbest{1.17}/\hlbest{1.78} & 2.07/\hlbest{1.70} &
    \hlsecond{1.32}/\hlbest{1.76} & \hlbest{0.88}/\hlbest{1.14} & \hlbest{1.42}/\hlbest{1.84} \\
    \bottomrule
    \end{tabular}}
\vspace{-2mm}
\end{table*}

%% file: figures/fig_compare_redwood.tex
\begin{figure*}[!t]
  \centering
  \includegraphics[width=.95\linewidth]{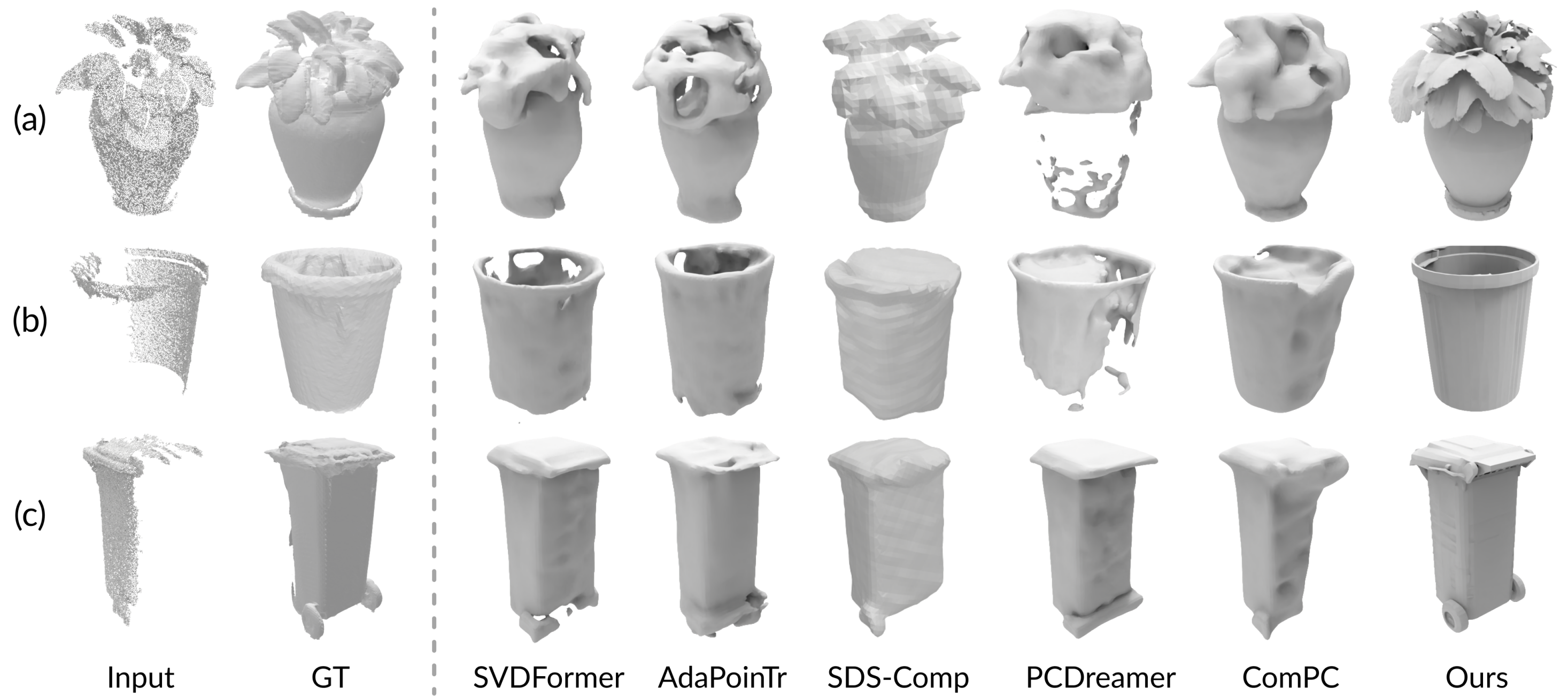}
  \vspace{-1mm}
  \caption{Qualitative comparison on Redwood dataset ~\cite{redwood}. We compare with various supervised and unsupervised methods \cite{zhu2023svdformer, adapointr, sdscomplete, pcdreamer, compc}, and visualize the output as meshes utilizing the commonly-used mesh reconstruction method~\cite{Peng2021SAP}.}
  \label{fig:viz_comp_redwood}
  \vspace{-3mm}
\end{figure*}

%% file: Tables/table_plyobj.tex
\begin{table*}[t]
\centering
\caption{
Quantitative comparisons on the synthetic data~\cite{compc, lipman2008green}.
We highlight the \bestcap{best}{} and \secondcap{second-best}{} results.
}
\vspace{-3mm}
\label{tab:exp_synthetic_cd_emd}
\setlength{\tabcolsep}{2pt}
\resizebox{\linewidth}{!}{
    \begin{tabular}{l|cccccccccccc|c}
    \toprule
    \textbf{CD $\downarrow$ / EMD $\downarrow$} & \textit{Horse} & \textit{Max-Planck} & \textit{Armadillo} & \textit{Cow} & \textit{Homer} & \textit{Teapot} & \textit{Bunny} & \textit{Nefertiti} & \textit{Bimba} & \textit{Ogre} & \textit{Lucy} & \textit{Dragon} & \textit{Average} \\
    \midrule
    SVDFormer~\cite{zhu2023svdformer} &
    4.33/5.17 & 8.84/7.95 & 4.97/6.13 & 3.50/4.39 &
    2.30/3.20 & 5.00/5.91 & 9.61/9.19 & 5.46/5.94 &
    7.54/7.18 & 4.90/5.55 & 2.08/2.93 & 2.88/4.35 & 5.12/5.66 \\
    
    AdaPoinTr~\cite{adapointr} &
    4.88/5.45 & 8.60/8.52 & 5.14/5.97 & 3.48/4.51 &
    2.28/3.30 & 3.92/4.53 & 9.33/8.86 & 5.54/6.16 &
    8.16/7.62 & 4.53/5.41 & 1.85/2.79 & 3.07/4.54 & 5.07/5.64 \\
    
    Shape-Inv~\cite{shapeinv} &
    6.55/10.4 & 4.94/5.72 & 4.79/7.22 & 4.74/7.83 & 2.36/3.75 &
    3.29/4.53 & 5.79/6.58 & 4.39/5.70 & 5.29/6.85 & 5.76/9.51 &
    2.99/4.07 & 4.48/7.91 & 4.62/6.68 \\
    
    PCDreamer~\cite{pcdreamer} &
    2.52/4.34 & 5.20/3.85 & 2.78/4.53 & 1.79/3.06 & 1.54/2.97 &
    1.95/3.57 & 5.48/5.66 & 2.72/3.43 & 3.83/5.39 & 2.57/3.97 &
    1.50/2.61 & 2.41/4.97 & 2.86/4.11 \\
    
    ComPC~\cite{compc} &
    1.29/1.77 & 1.22/\hlsecond{1.51} & 2.18/3.23 & 1.78/1.90 &
    1.32/1.65 & 1.09/1.34 & \hlbest{1.46}/\hlsecond{1.76} & 1.84/2.13 &
    1.65/1.90 & 1.56/2.11 & 1.97/2.70 & 1.99/3.10 & \hlsecond{1.61}/2.09 \\
    
    \midrule
    
    Ours (Direct3D-S2) &
    \hlbest{0.96}/\hlsecond{1.42} & \hlsecond{1.11}/1.63 & \hlsecond{1.15}/\hlsecond{1.67} & \hlbest{1.21}/\hlsecond{1.63} &
    \hlsecond{0.87}/\hlsecond{1.26} & \hlsecond{0.88}/\hlsecond{1.24} & \hlsecond{1.55}/1.97 & \hlsecond{0.96}/\hlsecond{1.40} &
    \hlbest{1.11}/\hlsecond{1.58} & \hlsecond{1.03}/\hlsecond{1.51} &
    \hlbest{0.96}/\hlbest{1.43} & \hlbest{1.51}/\hlbest{2.12} & \hlbest{1.11}/\hlsecond{1.57} \\
    
    Ours (TRELLIS) &
    \hlsecond{0.97}/\hlbest{1.31} & \hlbest{0.89}/\hlbest{1.20} & \hlbest{1.06}/\hlbest{1.48} & \hlsecond{1.23}/\hlbest{1.54} &
    \hlbest{0.62}/\hlbest{0.87} & \hlbest{0.61}/\hlbest{0.82} & 1.60/\hlbest{1.73} & \hlbest{0.81}/\hlbest{1.12} &
    \hlsecond{1.50}/\hlbest{1.55} & \hlbest{0.86}/\hlbest{1.16} &
    \hlsecond{1.30}/\hlsecond{1.87} & \hlsecond{1.73}/\hlsecond{2.36} & \hlbest{1.11}/\hlbest{1.41} \\
    \bottomrule
    \end{tabular}}
\vspace{-3mm}
\end{table*}

%% file: Tables/table_kitti_scannet.tex
\begin{table}[t]
\centering
\caption{
Completion fidelity on ScanNet~\cite{dataset_scannet} and KITTI~\cite{dataset_kitti}.
}
\vspace{-3mm}
\label{tab:exp_ucd_uhd_scannet_kitti}
\setlength{\tabcolsep}{5.8pt}
\resizebox{\linewidth}{!}{
    \begin{tabular}{l|ccc}
    \toprule
    \textbf{UCD $\downarrow$ / UHD $\downarrow$} & \textit{ScanNet-Chair} & \textit{ScanNet-Table} & \textit{KITTI-Car} \\
    \midrule
    SVDFormer~\cite{zhu2023svdformer} & 1.4 / 2.4 & 1.4 / 2.4 & 1.8 / 5.0 \\
    AdaPoinTr~\cite{adapointr}        & 1.4 / 2.4 & 1.3 / 2.4 & 1.6 / 4.9 \\
    Shape-Inv~\cite{shapeinv}         & 4.0 / 9.3 & 6.6 / 11.0 & 5.3 / 12.5 \\
    P2C~\cite{cui2023p2c}             & 3.4 / 5.0 & 2.4 / 4.9 & 3.9 / 5.8 \\
    PCDreamer~\cite{pcdreamer}        & 1.5 / 3.5 & 1.3 / 3.7 & 2.7 / 6.6 \\
    ComPC~\cite{compc}                & 2.0 / 5.3 & 3.0 / 7.0 & \textbf{1.1} / 5.7 \\
    \midrule
    Ours                    & \textbf{0.8} / \textbf{2.0} & \textbf{0.9} / \textbf{2.0} & 1.4 / \textbf{4.5} \\
    \bottomrule
    \end{tabular}}
    \vspace{-1mm}
\end{table}

%% file: Tables/table_omni_comp.tex
\begin{table}[t]
\centering
\caption{
Quantitative comparisons on our proposed Omni-Comp.
}
\vspace{-3mm}
\label{tab:cd_emd_three_patterns}
\resizebox{\linewidth}{!}{
    \begin{tabular}{l|ccc}
    \toprule
    \textbf{CD $\downarrow$ / EMD $\downarrow$} & \textit{Single Scan} & \textit{Random Crop} & \textit{Semantic Part} \\
    \midrule
    SVDFormer~\cite{zhu2023svdformer} & 5.32 / 6.84 & 5.35 / 6.73 & 5.66 / 7.19 \\
    AdaPoinTr~\cite{adapointr}        & 5.27 / 6.87 & 5.33 / 6.68 & 5.59 / 7.23 \\
    Shape-Inv~\cite{shapeinv}         & 4.98 / 7.31 & 5.39 / 6.79 & 5.37 / 6.91 \\
    SDS-Comp~\cite{sdscomplete}       & 5.42 / 6.27 & 5.12 / 6.35 & 5.60 / 6.75 \\
    PCDreamer~\cite{pcdreamer}        & 3.32 / 4.74 & 3.78 / 4.55 & 3.88 / 4.98 \\
    ComPC~\cite{compc}                & 4.24 / 4.61 & 5.48 / 5.95 & 6.37 / 6.18 \\
    \midrule
    Ours                     & \textbf{2.21} / \textbf{3.15} & \textbf{2.60} / \textbf{3.31} & \textbf{3.30} / \textbf{3.68} \\
    \bottomrule
    \end{tabular}}
    \vspace{-1mm}
\end{table}

%% file: figures/fig_scannet_kitti.tex
\begin{figure}[t]
    \centering
    \includegraphics[width=1.\linewidth]{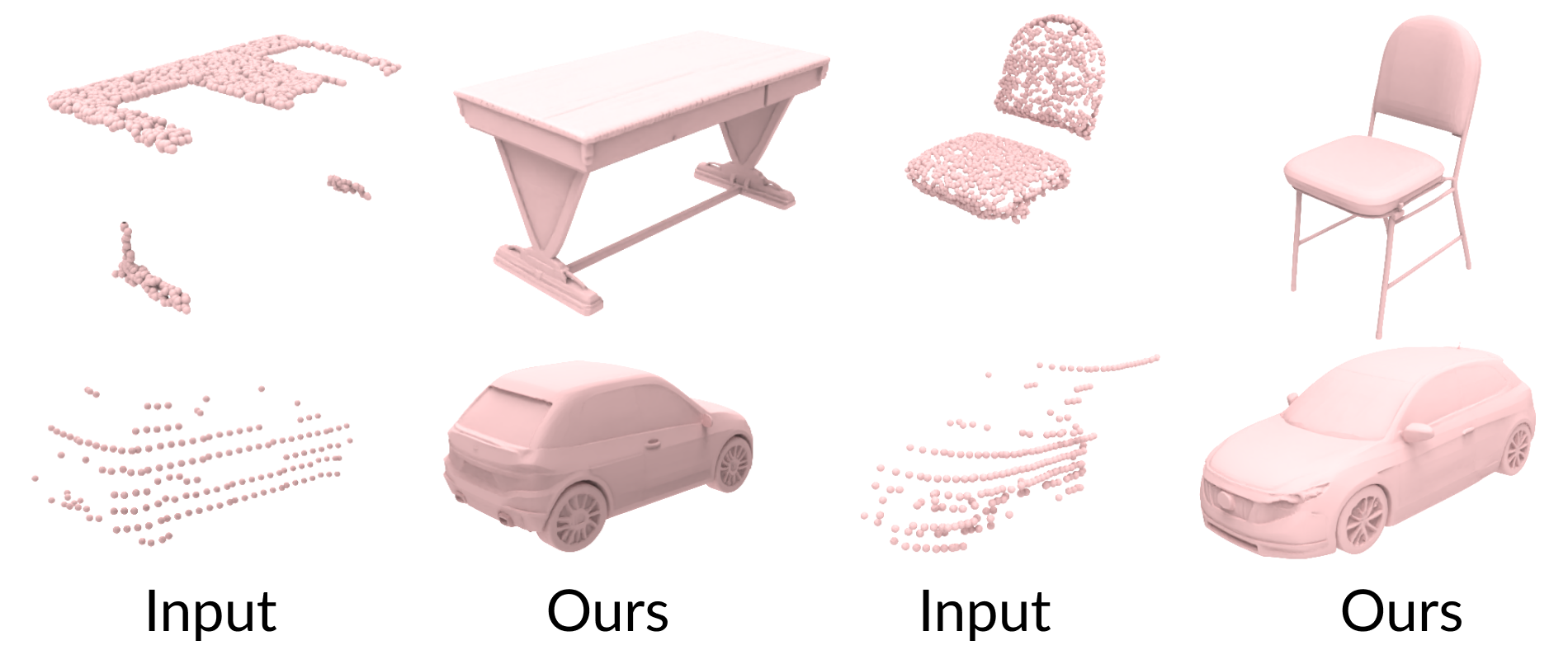}
    \vspace{-4mm}
    \caption{Visual examples on ScanNet \cite{dataset_scannet} and KITTI \cite{dataset_kitti} real-world datasets, which only contain real scans of table, chair, and car, with very sparse points.}
    \label{fig:scannet_and_kitti}
    \vspace{-5mm}
\end{figure}

%% file: figures/fig_compare_omni.tex
\begin{figure}[t]
  \centering
  \includegraphics[width=\linewidth]
  {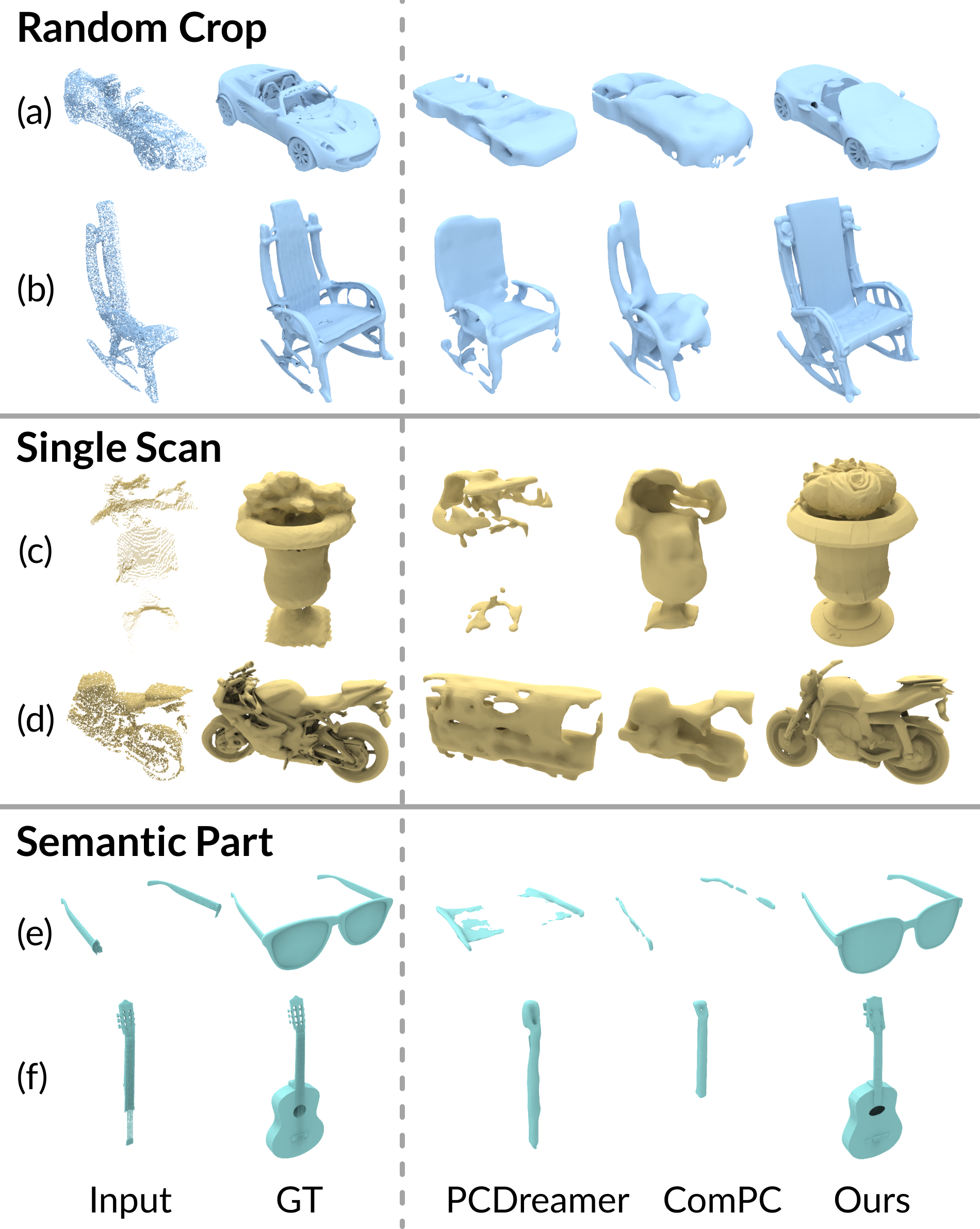}
  \vspace{-5mm}
  \caption{Qualitative comparisons on cases with different partial patterns from our Omni-Comp benchmark. Our approach produces more reasonable results than the latest methods~\cite{compc, pcdreamer}.}
  \label{fig:viz_comp_omni}
  \vspace{-3mm}
\end{figure}

%% file: Tables/table_mmd_tmd.tex
\begin{table}[t]
\centering
\caption{Completion diversity evaluation on Redwood~\cite{redwood} and synthetic data~\cite{compc, lipman2008green}.
}
\label{tab:mmd_tmd_redwood_synthetic}
\setlength{\tabcolsep}{20pt}
\resizebox{\linewidth}{!}{
    \begin{tabular}{l|c|c}
    \toprule
    \textbf{MMD $\downarrow$ / TMD $\uparrow$} & \textit{Redwood} & \textit{Synthetic} \\
    \midrule
    PCDreamer~\cite{pcdreamer} & 1.88 / 0.13 & 2.86 / 0.11 \\
    ComPC~\cite{compc}         & 2.01 / 0.72 & 1.65 / 0.62 \\
    \midrule
    Ours (Direct3D-S2)   & 1.81 / \textbf{1.35} & 1.23 / 0.89 \\
    Ours (TRELLIS)    & \textbf{1.62} / 0.99 & \textbf{1.20} / \textbf{0.94} \\
    \bottomrule
    \end{tabular}}
    \vspace{-2mm}
\end{table}

%% file: Tables/table_ablation.tex
\begin{table}[t]
\centering
\caption{Ablation studies. The baseline uses only latent replacement for completion.}
\label{tab:ablation_cd_emd}
\resizebox{\linewidth}{!}{
    \begin{tabular}{l|c|c}
    \toprule
    \textbf{CD $\downarrow$ / EMD $\downarrow$} & \textit{Redwood} & \textit{Synthetic} \\
    \midrule
    (a) Baseline (latent replacement) & 2.15 / 2.93 & 2.33 / 2.78 \\
    (b) Full w/o ERS                & 3.42 / 4.94 & 3.53 / 4.85 \\
    (c) Full w/o PNS                & 1.94 / 2.56 & 2.27 / 2.67 \\
    (d) Full w/o IAS                & 1.88 / 2.14 & 1.17 / 1.56 \\
    (e) IAS w/ 10 optimization steps  & 1.42 / 1.87 & 1.20 / 1.48 \\
    \midrule
    (f) Full pipeline (Ours) & \textbf{1.42} / \textbf{1.84} & \textbf{1.11} / \textbf{1.41} \\
    \bottomrule
    \end{tabular}}
    \vspace{-3mm}
\end{table}

%% file: figures/fig_diversity.tex
\begin{figure}[t]
    \vspace{0.cm}
    \centering
    \includegraphics[width=1.\linewidth]{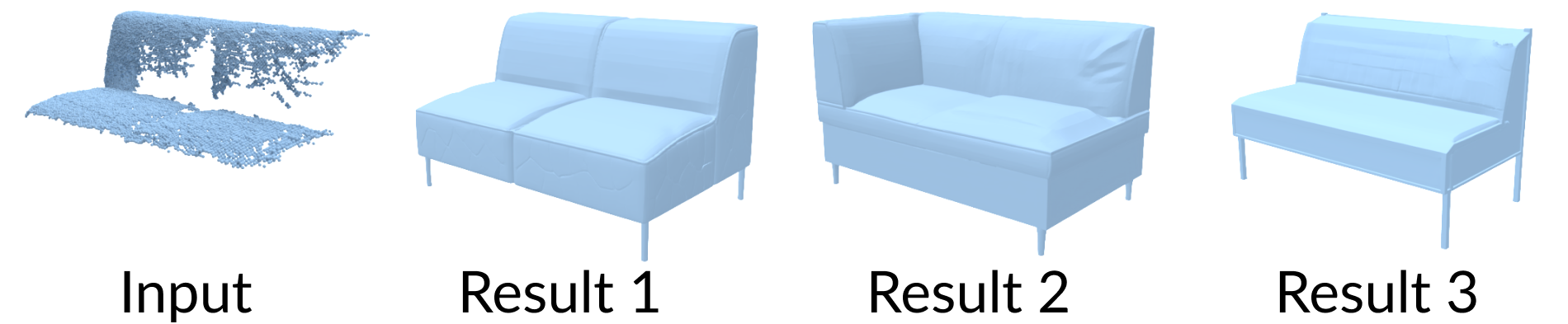}
    \vspace{-6mm}
    \caption{Visual examples of the completion diversity from our method on the real-world data.}
    \label{fig:diversity}
    \vspace{-3mm}
\end{figure}

%% file: figures/fig_ablation.tex
\begin{figure}[t]
    \centering
    \includegraphics[width=1.\linewidth]{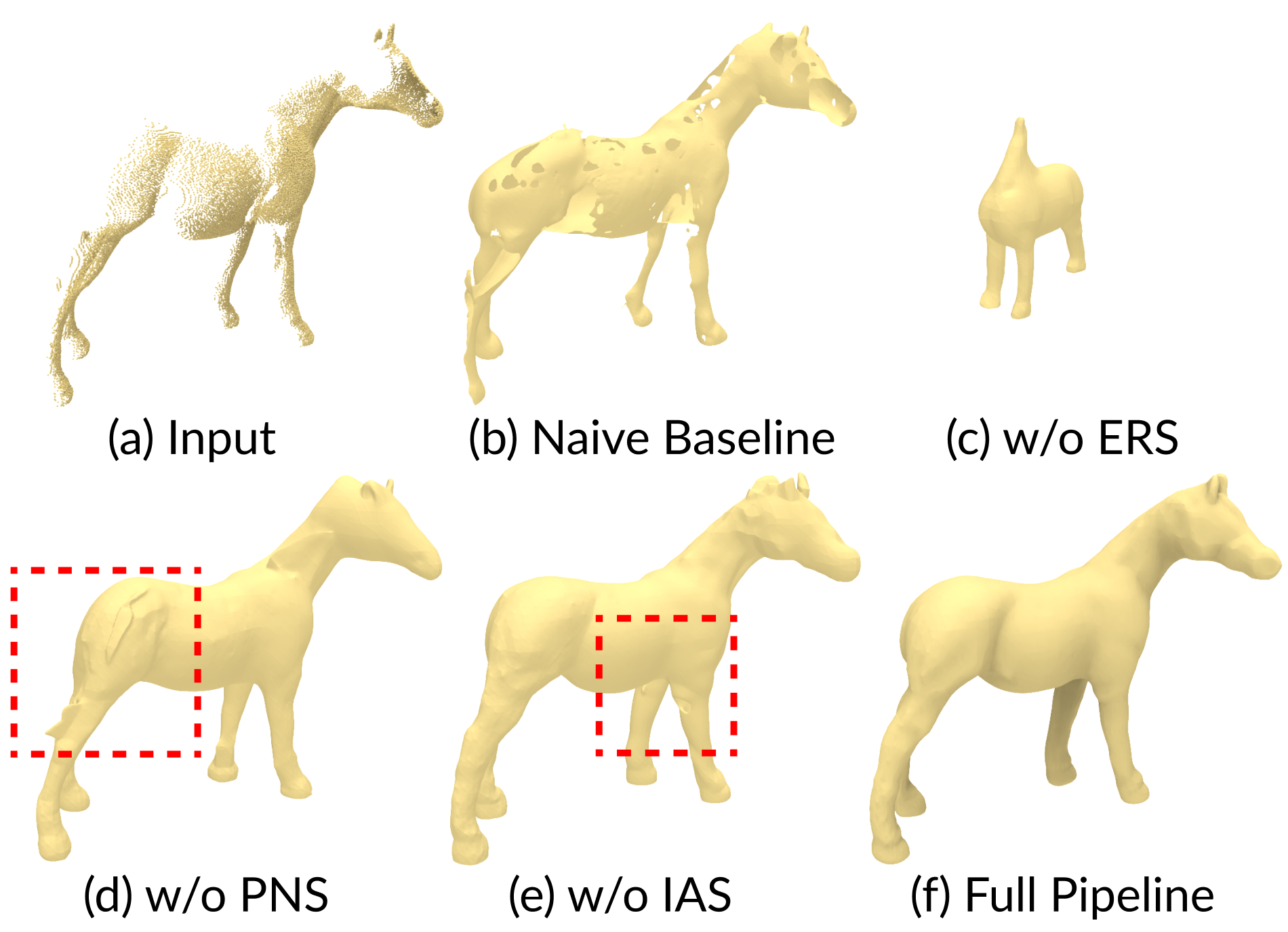}
    \vspace{-6mm}
    \caption{Visual comparison of the ablation studies. 
    The red boxes highlight the artifacts and holes.}
    \label{fig:ablation}
    \vspace{-3mm}
\end{figure}

%% file: sec/5_conclusion.tex
\section{Conclusion}
We presented LaS-Comp, a zero-shot and category-agnostic framework for 3D shape completion that leverages the rich geometric priors of 3D foundation models. The method integrates two complementary components, an Explicit Replacement Stage and an Implicit Alignment Stage, to jointly ensure high-fidelity reconstruction and global geometric coherence. 
We also introduce Omni-Comp, a benchmark combining real-world scans and synthetic shapes with diverse partial patterns for comprehensive evaluation. Experiments on Omni-Comp and standard benchmarks show that our method generalizes well across diverse categories and partial patterns, surpassing existing approaches.


\textbf{Acknoledgement}
This work was supported in part by the National Natural Science Foundation of China (NSFC) under grant 62372091 and in part by the Hainan Province Science and Technology Plan Project under Grant ZDYF2024(LALH)001.
\clearpage

%% file: Supp/X_suppl.tex
\clearpage
\onecolumn
\setcounter{page}{1}
\pagestyle{plain}

\begin{center}
    {\large\bfseries LaS-Comp: Zero-shot 3D Completion with Latent--Spatial Consistency \par}
    \vspace{0.5em}
    {\Large\itshape (Supplementary Material)}
\end{center}

\vspace{1.5em}

The supplementary material is organized as follows.
We discuss the latent–spatial gap between partial shapes and ground truth in \cref{sec:latent-spatial-gap}, detail the evaluation metrics in \cref{sec:metrics}, and provide additional implementation details in \cref{sec:implementation_details}.
\cref{sec:Omni-Comp} presents further details of the Omni-Comp benchmark.
Additional qualitative visualizations are shown in \cref{sec:more_viz}, and limitations and future directions are summarized in \cref{sec:limitation}.

\input{figures/fig_motivation}
\section{More Discussions about the Latent-Spatial Gap}
\label{sec:latent-spatial-gap}
As mentioned in the main paper, we observe that the latents of the partial input and the ground truth obtained by the VAE encoder in the observed regions differ significantly. As shown in \cref{fig:motivation}, we obtain the latents of $\boldsymbol{S}_{\text{p}}$ and $\boldsymbol{S}_{\text{gt}}$ respectively. Then we utilize the latent mask $\boldsymbol{M}$ downsampled from the voxel of $\boldsymbol{S}_{\text{p}}$ with resolution of $64^3$ to the latent space with resolution of $16^3$, to select the latents of both $\boldsymbol{S}_{\text{p}}$ and $\boldsymbol{S}_{\text{gt}}$ that belong to the partial regions, and compute the cosine similarity on the chosen latents. Computing such latent cosine similarity among all the samples with their GT on the Redwood dataset \cite{redwood}, we find that the average cosine similarity is only $\textbf{0.4593}$. 
%
%
As discussed in prior works~\cite{Evans1996, morris-etal-2020-reevaluating, yoo-qi-2021-towards-improving}, cosine similarities greater than 0.80 are generally regarded as reflecting strong correlations. In contrast, our measured average similarity of 0.4593 falls considerably below this threshold. This value indicates that the latent codes of the partial inputs and their corresponding GT shapes are far from strongly correlated in the overlapping regions.

This limited correlation suggests that directly relying on partial latent codes for completion guidance is fundamentally fragile, which is also verified by our ablation study in the main paper. Instead of performing completion in latent space, we project back to the original geometric space and explicitly inject the known partial points into the decoded space, so that the observed geometry is faithfully preserved and provides a stronger, more reliable constraint to guide the completion process.

\section{Details of the Evaluation Metrics}
\label{sec:metrics}
\noindent\textbf{Chamfer Distance (CD)} quantitatively evaluates the quality of completed point clouds, by adopting the symmetric distance between the predicted completion $\boldsymbol{S}_{\text{c}}$ and the ground-truth (GT) shape $\boldsymbol{S}_{\text{gt}}$. 
Let $\boldsymbol{S}_{\text{c}} = \{\boldsymbol{p}^{\text{c}}_i\}_{i=1}^{N} \subset \mathbb{R}^3$ denote the completed point cloud predicted by the model, and $\boldsymbol{S}_{\text{gt}} = \{\boldsymbol{p}^{\text{gt}}_j\}_{j=1}^{M} \subset \mathbb{R}^3$ denote the ground-truth point cloud. 
The Chamfer Distance is defined as the sum of two one-sided nearest-neighbor distances:
\begin{equation}
\label{eq:cd_def}
\mathrm{CD}(\boldsymbol{S}_{\text{c}}, \boldsymbol{S}_{\text{gt}})
=
\frac{1}{N} \sum_{i=1}^{N}
\min_{1 \leq j \leq M}
\bigl\|
\boldsymbol{p}^{\text{c}}_i - \boldsymbol{p}^{\text{gt}}_j
\bigr\|_2
+
\frac{1}{M} \sum_{j=1}^{M}
\min_{1 \leq i \leq N}
\bigl\|
\boldsymbol{p}^{\text{gt}}_j - \boldsymbol{p}^{\text{c}}_i
\bigr\|_2 .
\end{equation}
Here $\|\cdot\|_2$ denotes the Euclidean norm in $\mathbb{R}^3$.
The first term measures how well every predicted point is supported by the ground truth, while the second term measures how well the prediction covers the ground-truth surface. We report the metric in our experiments by $\times\! 10^{2}$.

\vspace{2mm}
\noindent\textbf{Earth Mover's Distance (EMD) \cite{emd2000}} is defined as the minimum average cost of transporting one set to the other.
After uniformly resampling of $\boldsymbol{S}_{\text{c}}$ and $\boldsymbol{S}_{\text{gt}}$, we assume $|\boldsymbol{S}_{\text{c}}| = |\boldsymbol{S}_{\text{gt}}| = N$.
The Earth Mover's Distance is defined as the minimum average cost of transporting one set to the other:
\begin{equation}
\label{eq:emd_def}
\mathrm{EMD}(\boldsymbol{S}_{\text{c}}, \boldsymbol{S}_{\text{gt}})
=
\min_{\phi:\,\boldsymbol{S}_{\text{c}} \to \boldsymbol{S}_{\text{gt}}}
\frac{1}{|\boldsymbol{S}_{\text{c}}|}
\sum_{\boldsymbol{p} \in \boldsymbol{S}_{\text{c}}}
\bigl\|
\boldsymbol{p} - \phi(\boldsymbol{p})
\bigr\|_2 ,
\end{equation}
where $\phi$ is a bijection between $\boldsymbol{S}_{\text{c}}$ and $\boldsymbol{S}_{\text{gt}}$.
This bijective matching is indicative of both the uniformity and the geometric fidelity of the generated shapes. Following ComPC \cite{compc}, we set eps as $0.005$, iteration as $50$ for the computation of EMD. We report the metric in our experiments by $\times\! 10^{2}$.

\vspace{2mm}
\noindent\textbf{Unidirectional Chamfer Distance (UCD)}  measures the squared $L_2$ distance from the partial input shape $\boldsymbol{S}_{\text{p}}$ to the completed output $\boldsymbol{S}_{\text{c}}$. 
Given the partial point set $\boldsymbol{S}_{\text{p}} \subset \mathbb{R}^3$ and the completed point set $\boldsymbol{S}_{\text{c}} \subset \mathbb{R}^3$, we define
\begin{equation}
\label{eq:ucd_def}
\mathrm{UCD}(\boldsymbol{S}_{\text{p}}, \boldsymbol{S}_{\text{c}})
=
\frac{1}{|\boldsymbol{S}_{\text{p}}|}
\sum_{\boldsymbol{p} \in \boldsymbol{S}_{\text{p}}}
\min_{\boldsymbol{q} \in \boldsymbol{S}_{\text{c}}}
\bigl\|
\boldsymbol{p} - \boldsymbol{q}
\bigr\|_2^2 .
\end{equation}
Unlike the symmetric Chamfer Distance, UCD only measures how well the completed shape preserves and explains the observed partial input. We report the metric in our experiments by $\times\! 10^{4}$.

\vspace{2mm}
\noindent\textbf{Unidirectional Hausdorff Distance (UHD) \cite{chen2019unpaired, wu2020multimodal}} similarly measures the single-sided Hausdorff distance from the partial input shape $\boldsymbol{S}_{\text{p}}$ to the completed output $\boldsymbol{S}_{\text{c}}$:
\begin{equation}
\label{eq:uhd_def}
\mathrm{UHD}(\boldsymbol{S}_{\text{p}}, \boldsymbol{S}_{\text{c}})
=
\max_{\boldsymbol{p} \in \boldsymbol{S}_{\text{p}}}
\min_{\boldsymbol{q} \in \boldsymbol{S}_{\text{c}}}
\bigl\|
\boldsymbol{p} - \boldsymbol{q}
\bigr\|_2 .
\end{equation}
Compared with UCD, UHD focuses on the worst-case (hardest-to-match) observed point in $\boldsymbol{S}_{\text{p}}$. We report the metric in our experiments by $\times\! 10^{2}$.

\vspace{2mm}
\noindent\textbf{Minimum Matching Distance (MMD) \cite{mmd}} measures the fidelity of a set of generated completed shapes with respect to the set of ground-truth shapes.
Let $\mathcal{D}_{\text{c}} = \{\boldsymbol{S}_{\text{c}}^{(i)}\}_{i=1}^{N_{\text{c}}}$ be the set of completed shapes and
$\mathcal{D}_{\text{gt}} = \{\boldsymbol{S}_{\text{gt}}^{(j)}\}_{j=1}^{N_{\text{gt}}}$ be the set of ground-truth shapes, where each
$\boldsymbol{S}_{\text{c}}^{(i)}$ and $\boldsymbol{S}_{\text{gt}}^{(j)}$ is a point cloud in $\mathbb{R}^3$.
Given a point-set distance $\mathrm{CD}(\cdot,\cdot)$, the Minimum Matching Distance is defined as
\begin{equation}
\label{eq:mmd_def}
\mathrm{MMD}(\mathcal{D}_{\text{c}}, \mathcal{D}_{\text{gt}})
=
\frac{1}{|\mathcal{D}_{\text{gt}}|}
\sum_{\boldsymbol{S}_{\text{gt}} \in \mathcal{D}_{\text{gt}}}
\;
\min_{\boldsymbol{S}_{\text{c}} \in \mathcal{D}_{\text{c}}}
\mathrm{CD}\bigl(\boldsymbol{S}_{\text{gt}}, \boldsymbol{S}_{\text{c}}\bigr).
\end{equation}
That is, for each ground-truth shape we find the closest generated completion in $\mathcal{D}_{\text{c}}$ under $\mathrm{CD}(\cdot,\cdot)$ and report the average matched distance. We report the metric in our experiments by $\times\! 10^{2}$.

\vspace{2mm}
\noindent\textbf{Total Mutual Difference (TMD)} measures the diversity of multiple completion results generated for the same partial input.
Given a partial shape $\boldsymbol{S}_{\text{p}}$, we generate a set of $K$ completed shapes
$\mathcal{D}_{\text{c}} = \{\boldsymbol{S}_{\text{c}}^{(k)}\}_{k=1}^{K}$, where each $\boldsymbol{S}_{\text{c}}^{(k)}$ is a point cloud in $\mathbb{R}^3$.
We apply the Chamfer Distance defined in Eq.~\eqref{eq:cd_def} to assess the difference of each pair of point clouds inside $\mathcal{D}_{\text{c}}$.
The Total Mutual Difference for this sample is defined as the average pairwise distance among all completions:
\begin{equation}
\label{eq:tmd_def}
\mathrm{TMD}(\mathcal{D}_{\text{c}})
=
\frac{2}{K(K-1)}
\sum_{1 \leq a < b \leq K}
\mathrm{CD}\bigl(\boldsymbol{S}_{\text{c}}^{(a)}, \boldsymbol{S}_{\text{c}}^{(b)}\bigr).
\end{equation}
A larger TMD value indicates higher diversity among the generated completion hypotheses for the same partial input $\boldsymbol{S}_{\text{p}}$. We report the metric in our experiments by $\times\! 10^{2}$.

\section{More Implementation Details}
\label{sec:implementation_details}

\subsection{Backbones for 3D Completion}

Our framework is compatible with backbones that expose three abstract components: a spatial-to-latent encoder ($\mathcal{E}$), a latent-to-spatial decoder ($\mathcal{D}$), and a latent generative process ($\mathcal{G}$). We instantiate this with Direct3D-S2~\cite{wu2025direct3ds2gigascale3dgeneration} and TRELLIS~\cite{xiang2025structured}. For Direct3D-S2, which operates on the dense voxel grid, $\mathcal{E}$ is its VAE encoder, $\mathcal{G}$ is the conditional diffusion transformer, and $\mathcal{D}$ is the VAE decoder. For TRELLIS, which uses models for the first stage (sparse structure generation), $\mathcal{E}$ is the VAE Encoder, $\mathcal{G}$ is the first stage of its rectified-flow transformer generator, and $\mathcal{D}$ is the VAE Decoder. 
For both backbones, the resolution of the voxel grid is $64^3$; all the input to the models should be voxelized first. Our pipeline operates on the first stage of both backbones, which operates to generate dense voxel grids as the main geometry, while freezing the second stage of the models to generate SDF for meshes and sampling as point clouds for evaluation.

\subsection{Pipeline Parameter Settings}
In our experiments, we set the number of denoising time steps to $100$, the CFG scale to $1.0$, and the rescale factor of $t$ to $3.0$. At each time step, we perform one-step latent optimization update with a learning rate of $1\times 10^{-5}$.  The occupancy threshold is set to $0.5$, following the backbone settings~\cite{wu2025direct3ds2gigascale3dgeneration, xiang2025structured}. In the IAS stage, the output of the decoder $\mathcal{D}$ is kept as logits rather than thresholded into an occupancy grid; we directly compute the binary cross-entropy loss on these logits and optimize the latent. Our PNS with new sampling noise is applied only for $t \in [0.5, 1]$, while ERS and IAS are applied for all time steps. 

\subsection{Dataset Parameter Settings}
Following prior experimental settings~\cite{shapeinv, compc, cui2023p2c, chen2019unpaired}, we set the point cloud resolution of the Redwood~\cite{redwood} and synthetic~\cite{compc, lipman2008green} datasets to $16{,}384$, and that of KITTI~\cite{dataset_kitti} and ScanNet~\cite{dataset_scannet} to $2{,}048$. Our proposed Omni-Comp benchmark also uses a resolution of $16{,}384$.

\subsection{ 
Work on consumer-grade GPUs.} 
%
%
Our method can be implemented on a single RTX3090: under FP16 with batch size 1, it uses 10.29 GB peak VRAM and runs in 35.92s, which demonstrates the practical deployability.

\section{Details of the Omni-Comp Benchmark}
\label{sec:Omni-Comp}

We introduce \textbf{Omni-Comp}, a new benchmark designed for a more comprehensive and robust evaluation of 3D shape completion.
Our benchmark features a challenging set of $30$ objects, each from a distinct category (listed in \cref{tab:omnicomp_categories}), curated from diverse sources: $10$ real-world scans from Redwood~\cite{redwood} (chosen for complex geometry), $10$ real-world everyday objects from YCB~\cite{ycb} (motivated by downstream applications, \eg, robotic grasping), and $10$ synthetic shapes from \cite{partnextnext} (chosen for rich semantic structure, with complex geometry).
Critically, inspired by \cite{shapeinv}, our benchmark generates three distinct partial patterns for each object:
(i) \textit{Single Scan}: using the projection of the captured depth map for real-world data, and simulating a standard depth camera capture for synthetic data; 
(ii) \textit{Random Crop}: Representing arbitrary occlusions by randomly cropping a portion; and 
(iii) \textit{Semantic Part}: Keeping a semantic component and removing other parts. 
By creating two samples for each pattern per object, the benchmark comprises $180$ challenging partial samples with corresponding ground truths. 

When selecting real-world data from \cite{redwood, ycb}, we intentionally avoid near-cuboidal or purely box-like geometries and instead prioritize objects with richer, more complex structures. For the synthetic data, we choose samples whose identities do not appear in the released Objaverse-XL \cite{deitke2023objaversexl} metadata or index, so that they are not included in the training data of our generative backbones.

We introduce the preprocessing of samples for the three partial patterns as follows. \textbf{(a)} When constructing samples of a single scan, if not providing the object mask, we utilize SAM2 \cite{sam2} and OWL-ViT \cite{owlvit} to extract the mask in the real-world RGB-D data, where the depth map is registered to the image. After that, we back-project the depth according to the mask to the 3D space. Next, we follow SDS-Comp \cite{sdscomplete}, manually align the provided ground truth point clouds with the back-projected partial point cloud, and apply ICP for refinement \cite{icp}. The ground truth with the aligned partial scan is normalized by the bounding box of GT, to the range of $[-0.5, 0.5]$. For the synthetic data, we establish a virtual camera with virtual scans to construct the partial point clouds.
\textbf{(b)} Considering the random crop, we sample axis-aligned slabs and half-space cuts along random axes. Concretely, we select percentile-based windows or half-spaces along a chosen axis to retain only a random fraction of the GT object (e.g., $40$-$70\%$), so that each crop mimics realistic partial views where only a contiguous portion of the object is observed. 
\textbf{(c)}
Speaking of the semantic part pattern, we manually isolate meaningful object parts on the complete object and remove all remaining geometry using an interactive editing tool. This yields semantic partial shapes that consist of a single, coherent part, providing a distinct partial pattern compared with the single scan and random crop patterns. 

\begin{figure}[hb]
    \centering
    \includegraphics[width=1.0\linewidth]{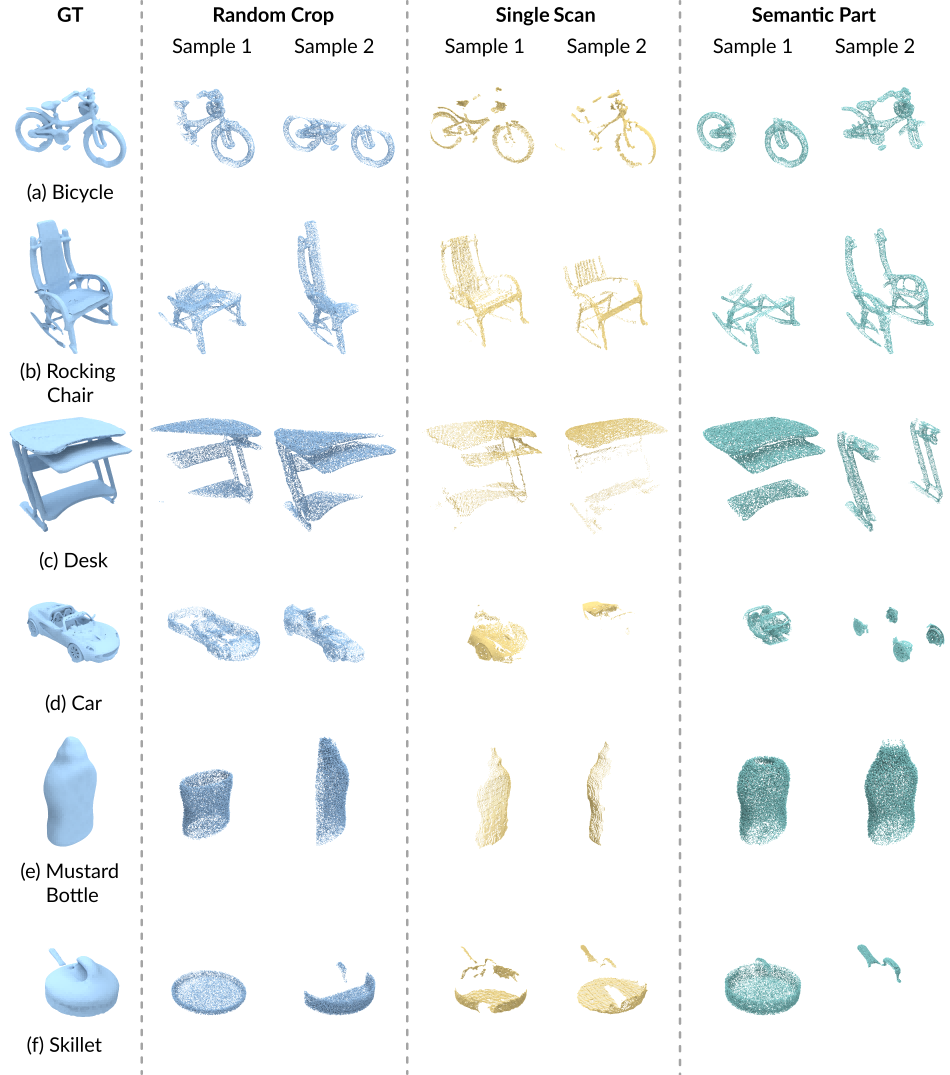}
    \caption{Visual examples of the proposed Omni-Comp benchmark. We show the ground truth mesh, with its corresponding partial samples of random crop, single scan, and semantic part.}
    \label{fig:supp_viz_omni_comp}
\end{figure}

\begin{figure}[t]
    \centering
    \includegraphics[width=1.0\linewidth]{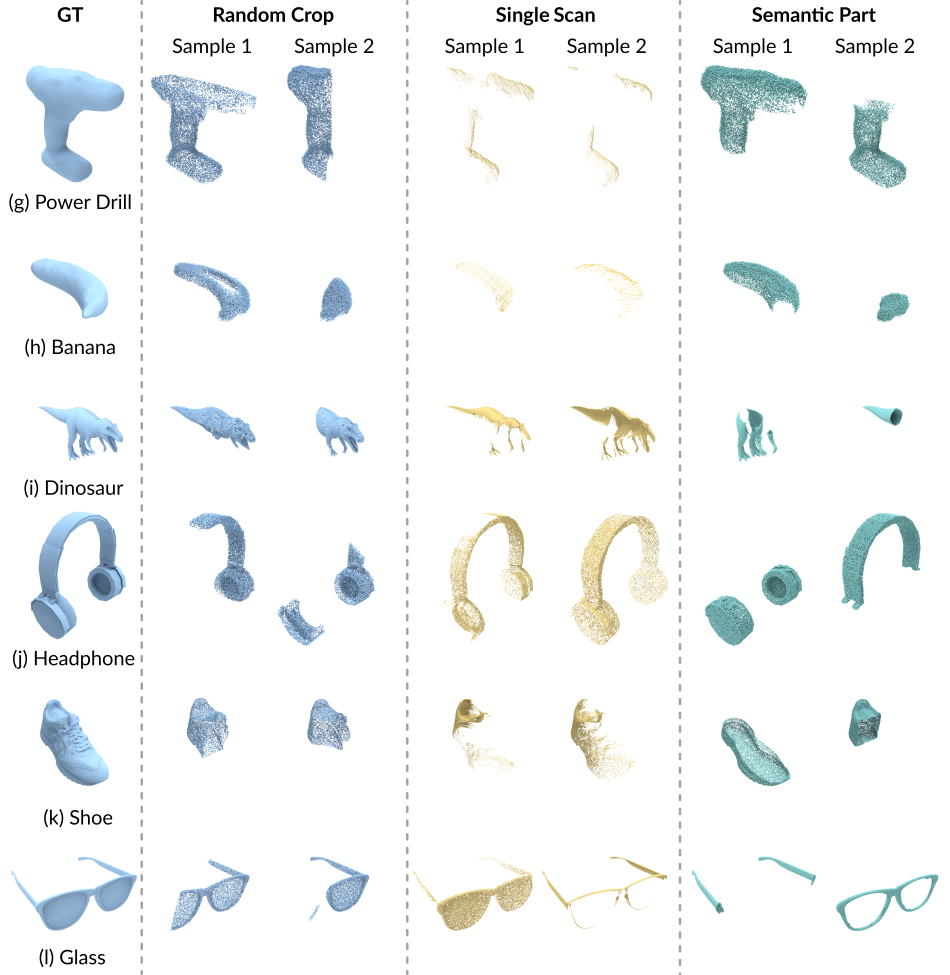}
    \caption{Visual examples of the proposed Omni-Comp benchmark.  We show the ground truth mesh, with its corresponding partial samples of random crop, single scan, and semantic part.}
    \label{fig:supp_viz_omni_comp_2}
\end{figure}

Note that all the partial and GT data are sampled with a resolution of $16{,}384$, following \cite{sdscomplete, compc}. We provide more visualizations of our dataset across diverse objects and different partial patterns, as shown in \cref{fig:supp_viz_omni_comp} and \cref{fig:supp_viz_omni_comp_2}.

%

\begin{table}[t]
\centering
\caption{
\textbf{Category list in the Omni-Comp benchmark.}
Representative object categories from Redwood~\cite{redwood},
YCB~\cite{ycb}, and Synthetic~\cite{partnextnext}.
}
\label{tab:omnicomp_categories}
\setlength{\tabcolsep}{5pt}
\resizebox{0.5\linewidth}{!}{
\begin{tabular}{c|l|c|l|c|l}
\toprule
\multicolumn{2}{c|}{\textit{Redwood}} &
\multicolumn{2}{c|}{\textit{YCB}} &
\multicolumn{2}{c}{\textit{Synthetic}} \\
\midrule
\textbf{ID} & \textbf{Category} &
\textbf{ID} & \textbf{Category} &
\textbf{ID} & \textbf{Category} \\
\midrule
1  & Bicycle         & 11 & Cracker Box        & 21 & Dinosaur \\
2  & Plant           & 12 & Mustard Bottle     & 22 & Glass \\
3  & Trash Bin       & 13 & Banana             & 23 & Guitar \\
4  & Car             & 14 & Bleach Cleanser    & 24 & Headphone \\
5  & Motorcycle      & 15 & Skillet            & 25 & Laptop \\
6  & Sign            & 16 & Power Drill        & 26 & Robot \\
7  & Bench           & 17 & Extra Large Clamp  & 27 & Shoe \\
8  & Couch           & 18 & Tennis Ball        & 28 & Statue \\
9  & Desk            & 19 & Wood Block         & 29 & Toilet \\
10 & Rocking Chair   & 20 & Timer              & 30 & Torch \\
\bottomrule
\end{tabular}}
\end{table}

\section{More Visualization Results}
\label{sec:more_viz}

\subsection{Visualization Results on the Redwood Dataset}
\begin{figure}[t]
    \centering
    \includegraphics[width=1.0\linewidth]{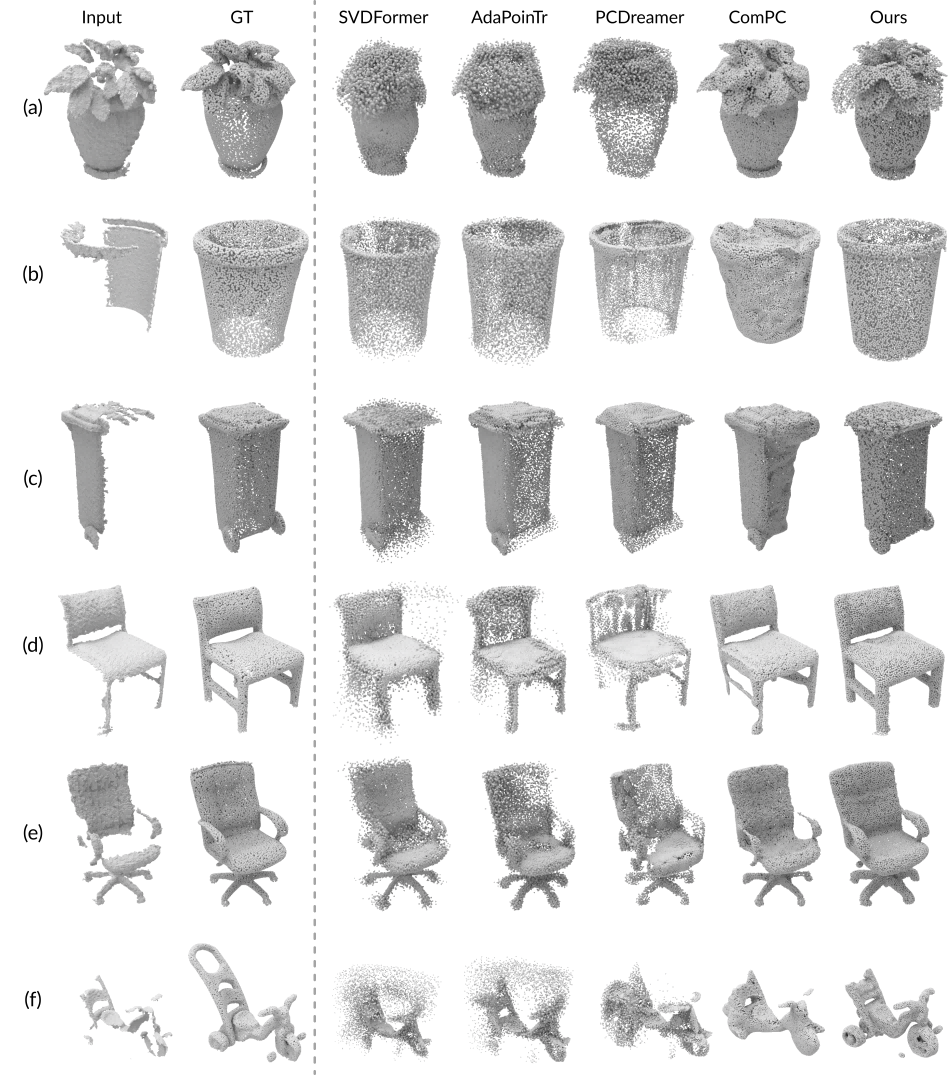}
    \caption{Visual Comparison on the Redwood dataset \cite{redwood}, with point cloud representation. Obviously, our method significantly outperforms prior approaches in most of the examples, with very high-quality geometry and good completion correctness.}
    \label{fig:supp_viz_redwood}
\end{figure}

To ensure a better understanding of our method's superior performance and a fair comparison setting, we provide more examples from the Redwood dataset \cite{redwood}, visualized as point clouds, in \cref{fig:supp_viz_redwood}. 
These real-world scans are highly challenging, containing strong sensor noise, self-occlusions, and depth discontinuities. 
Across a wide range of object instances and viewpoints, our method consistently recovers geometrically detailed and semantically plausible completions: the global structure of the object is preserved, while fine-scale parts such as legs, handles, and support structures are reasonably completed.

\subsection{Visualization Results on the Synthetic Dataset}
\begin{figure}[t]
    \centering
    \includegraphics[width=1.0\linewidth]{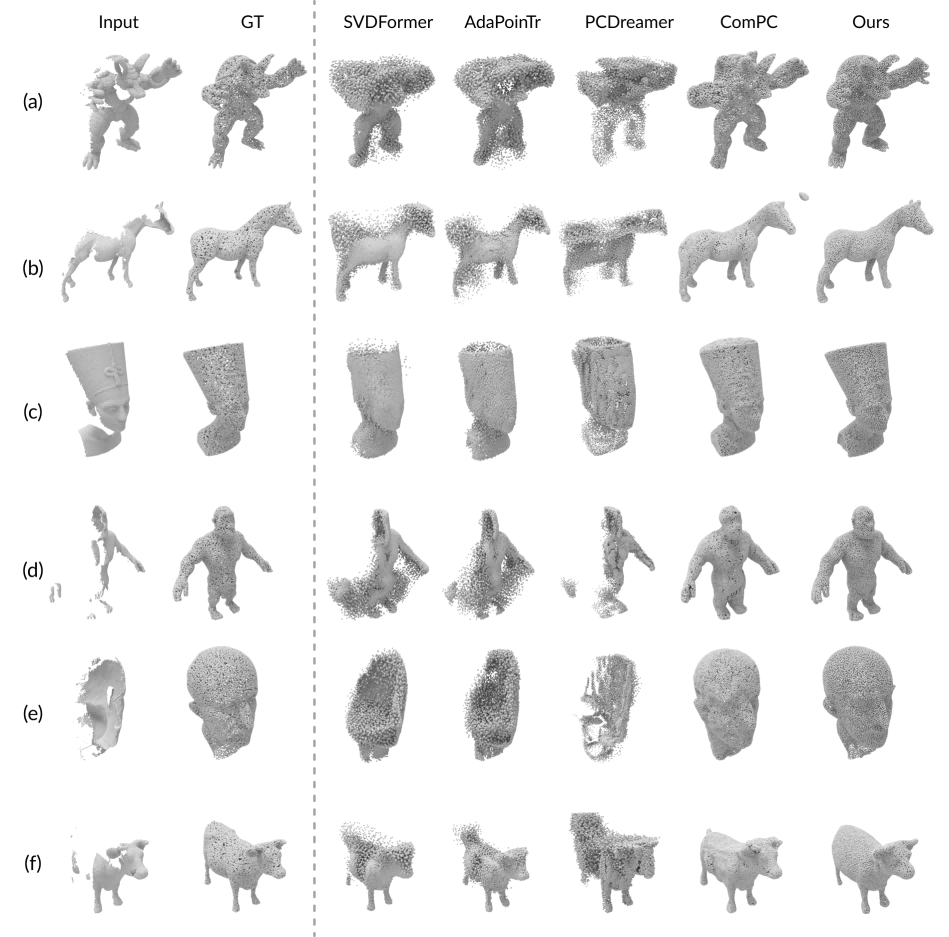}
    \caption{Visual Comparison on the synthetic dataset \cite{compc}, with point cloud representation. Obviously, our method significantly outperforms prior approaches in most of the examples, with very high-quality geometry and good completion correctness.}
    \label{fig:supp_viz_synthetic}
\end{figure}

We also provide qualitative examples on synthetic data from \cite{compc}, as visualized in \cref{fig:supp_viz_synthetic}. 
Compared with real-world scans, these synthetic partial point clouds exhibit cleaner sampling and more diverse geometric patterns, including thin structures, high-curvature regions, and complex topologies. 
Our method is able to accurately complete these partial observations into full shapes with rich geometric details and coherent semantics, showing that the proposed pipeline generalizes well to different domains of data.

\subsection{Visualization Results on the Proposed Omni-Comp Benchmark}
\begin{figure}[t]
    \centering
    \includegraphics[width=1.0\linewidth]{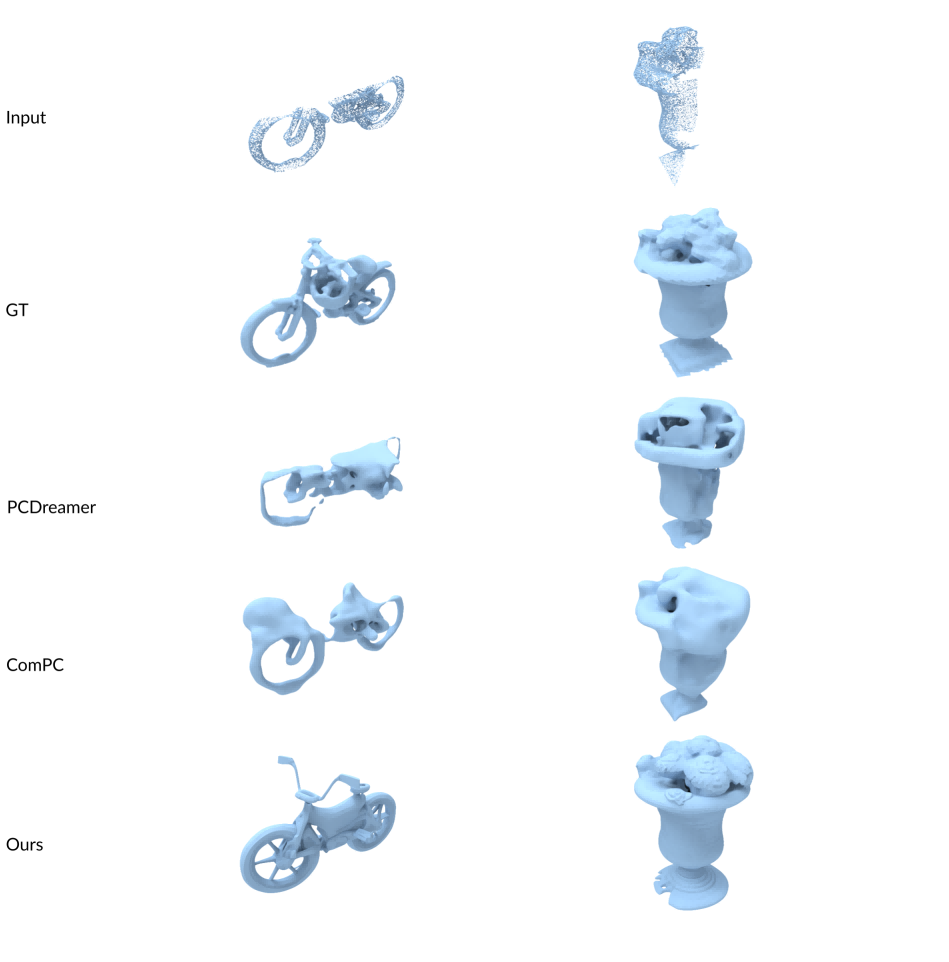}
    \caption{More visual examples on the proposed Omni-Comp benchmark, under the random crop partial pattern. Our method consistently provides better completion results compared with the latest methods \cite{pcdreamer, compc}, regardless of the partial pattern and object category.}
    \label{fig:supp_viz_omni_comp_random_crop}
\end{figure}

\begin{figure}[t]
    \centering
    \includegraphics[width=1.0\linewidth]{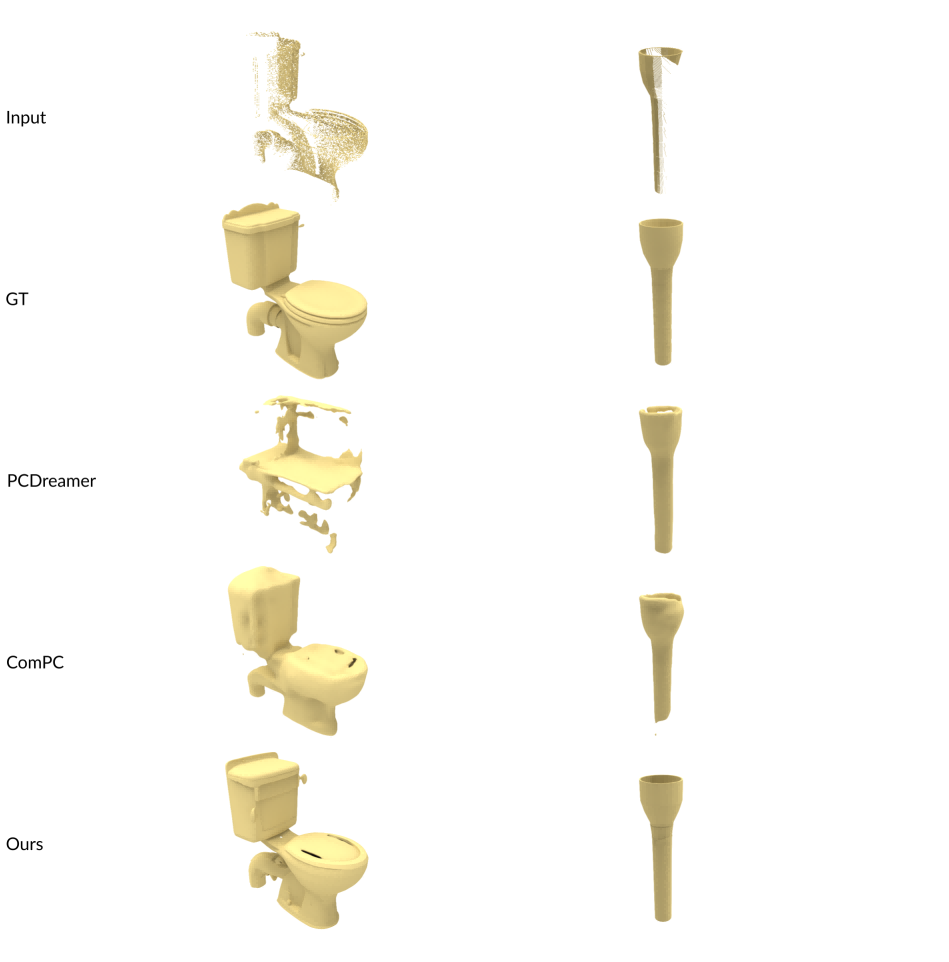}
    \caption{More visual examples on the proposed Omni-Comp benchmark, under the single scan partial pattern. Our method consistently provides better completion results compared with the latest methods \cite{pcdreamer, compc}, regardless of the partial pattern and object category.}
    \label{fig:supp_viz_omni_comp_single_scan}
\end{figure}

\begin{figure}[t]
    \centering
    \includegraphics[width=1.0\linewidth]{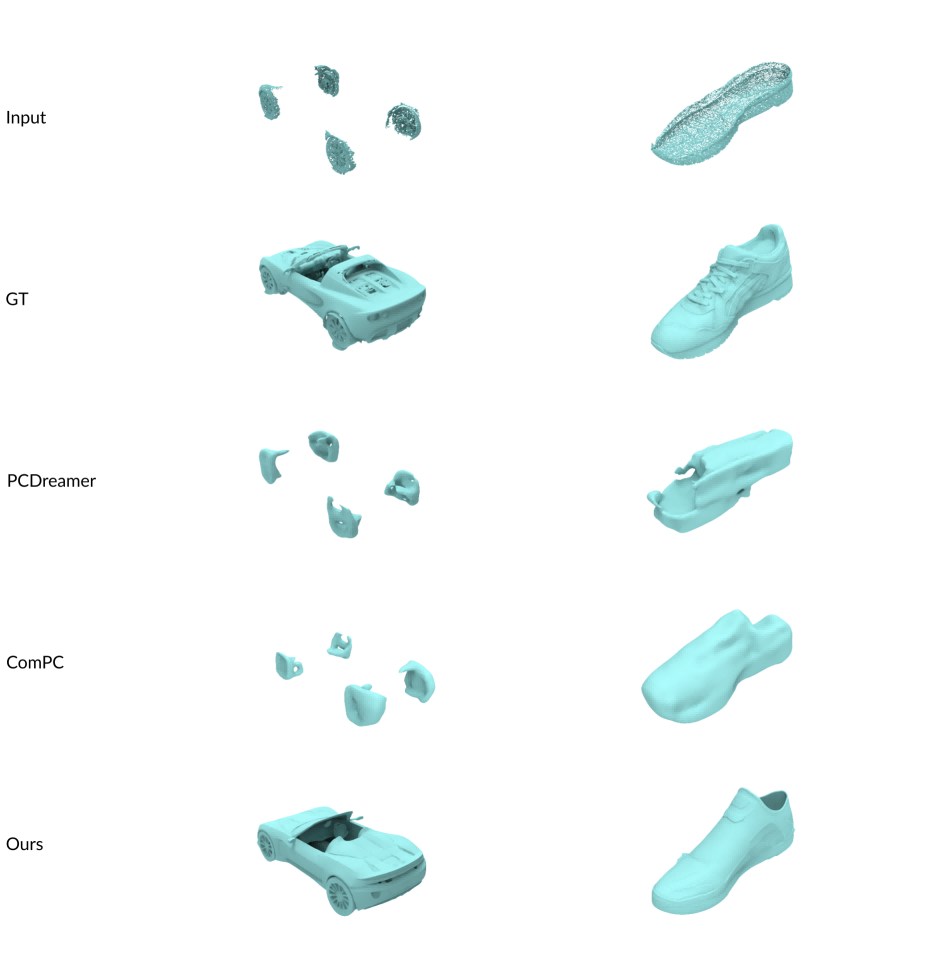}
    \caption{More visual examples on the proposed Omni-Comp benchmark, under the semantic part partial pattern. Our method consistently provides better completion results compared with the latest methods \cite{pcdreamer, compc}, regardless of the partial pattern and object category.}
    \label{fig:supp_viz_omni_comp_semantic_part}
\end{figure}

To further demonstrate our method's generalization ability across diverse partial patterns and object categories, we provide additional qualitative examples on our Omni-Comp benchmark in \cref{fig:supp_viz_omni_comp_random_crop}, \cref{fig:supp_viz_omni_comp_single_scan}, and 
\cref{fig:supp_viz_omni_comp_semantic_part}, respectively. 
The benchmark covers multiple partial patterns (random crop, single scan, and semantic part) and spans a wide spectrum of categories. 
Under all these settings, our method produces geometrically detailed and semantically meaningful completions that respect the observed regions while plausibly hallucinating the missing parts. 
These results highlight the strong robustness and cross-pattern generalization of our framework when faced with heterogeneous partial inputs.

\subsection{Visualization Results on Completion Diversity}
\begin{figure}[t]
    \centering
    \includegraphics[width=1.0\linewidth]{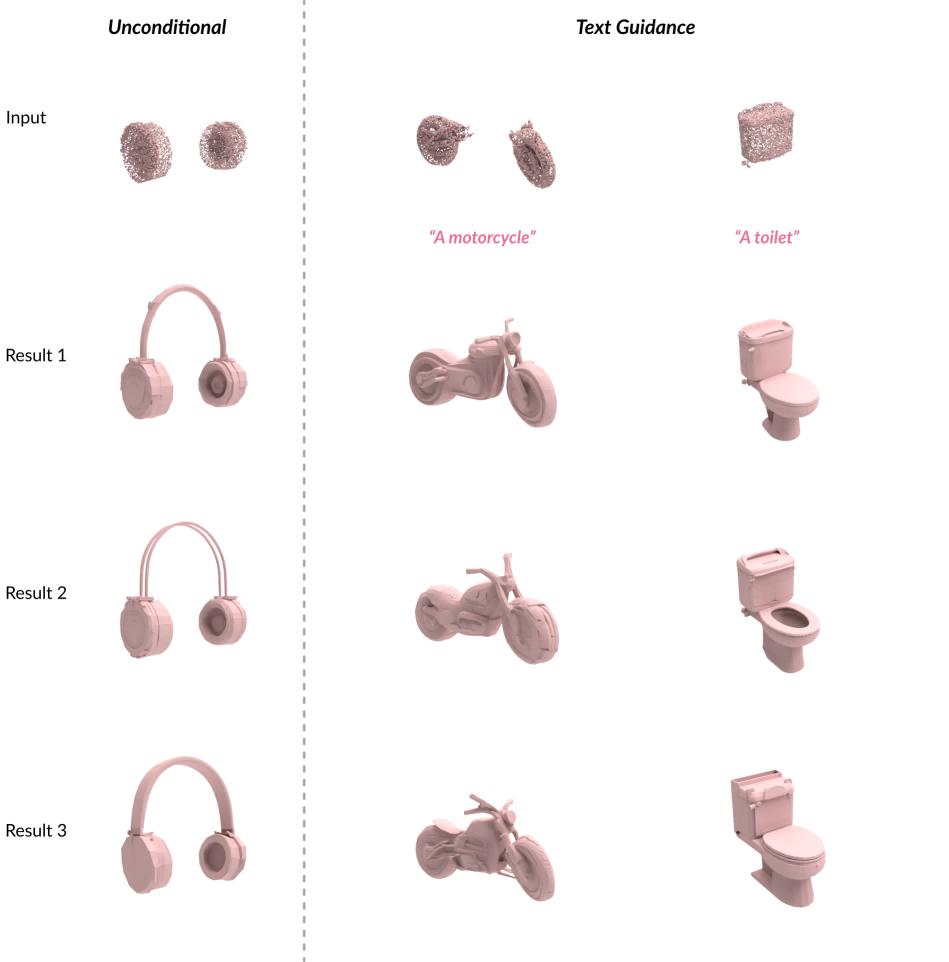}
    \caption{More visual examples to show the completion diversity, under unconditional and text-guided completion. In both completion settings, our method can provide reasonable results with good geometry.}
    \label{fig:supp_viz_diversity_uncond_text}
\end{figure}

Finally, to verify that our method can generate diverse yet reasonable completions for the same partial input, we present additional qualitative examples in \cref{fig:supp_viz_diversity_uncond_text} under both unconditional completion and text-guided completion. 
Given a fixed partial point cloud, our model is able to produce multiple distinct full-shape hypotheses that all remain consistent with the observed geometry, illustrating its ability to capture the inherent ambiguity of the completion task. 
These visualizations demonstrate that our approach not only yields high-quality and geometrically refined completions but also supports meaningful multimodal diversity in the output space.

\begin{figure}[htb]
    \centering
    \vspace{-3mm}
    \includegraphics[width=1.0\linewidth]{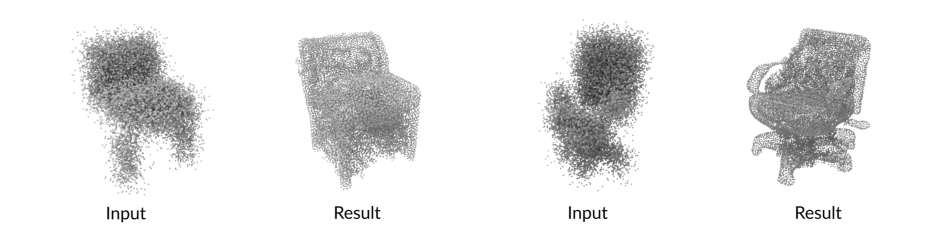}
    \caption{Visual examples of the completion results under extremely noisy partial inputs. Despite the severe noise that heavily corrupts the observed points, our method can still recover a reasonable global structure and overall object silhouette, but many fine details and thin structures are degraded or missing, revealing the limitation of our model when strong noise overwhelms the underlying geometry.}
    \label{fig:supp_viz_noise}
    \vspace{-3mm}
\end{figure}

\section{Limitations and Future Work}
\label{sec:limitation}
Although our method outperforms existing approaches in various benchmarks, object categories, and partial patterns, extremely noisy inputs remain challenging and may still lead to imperfect completions; see \cref{fig:supp_viz_noise}.
In such cases, the model can still recover the coarse object structure,~\eg, overall chair silhouette, but fine details and thin structures may be over-smoothed or distorted because the underlying geometry is heavily corrupted. When the input contains only weak or ambiguous cues, the generative prior has limited reliable information to condition on.
To further improve robustness in such scenarios, our future work will explore:
a) stronger outlier-removal designs that explicitly detect and remove noise patterns such as scattered points far from the main shape or clusters with inconsistent local geometry before completion, thereby exposing a cleaner partial input for shape completion; and b) confidence-aware refinement strategies that adapt the denoising strength based on the model’s estimated reliability in each region, preserving confident areas while applying more cautious updates to ambiguous or noisy regions.

What's more, following ComPC, SDS-Comp, and GenPC, we adopt GT-based normalization. 
However, for real-world application and implementation, we usually encounter situations where only partial scans can be obtained, while the gt shapes are missing. In the future, we plan to develop more robust methods for such in-the-wild 3D completion.

%% file: figures/fig_motivation.tex
\begin{figure}[htb]
    \centering
    \includegraphics[width=0.6\linewidth]{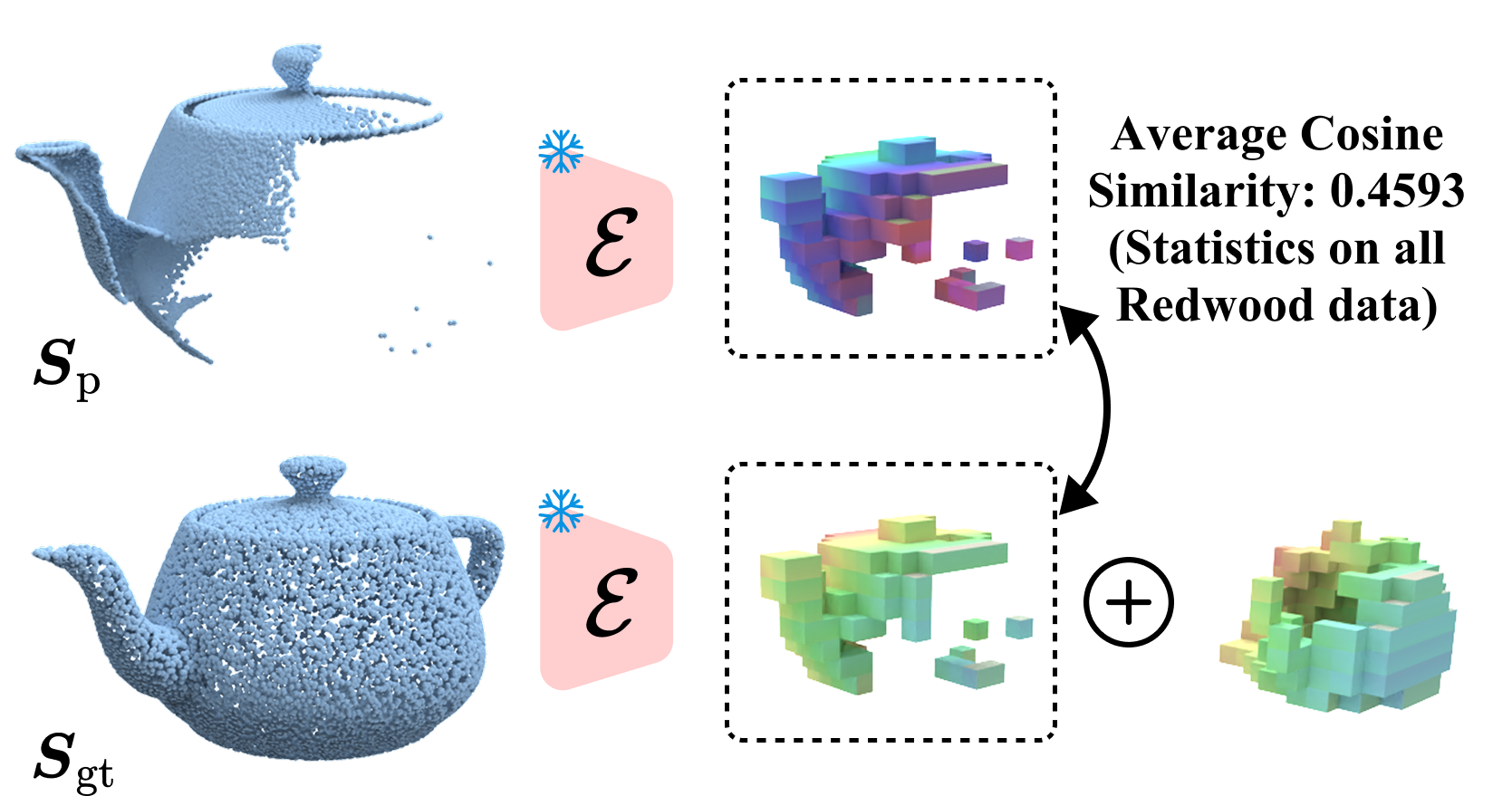}
    \caption{Illustration of the latent gap between the partial input $\boldsymbol{S}_{\text{p}}$ and the ground truth $\boldsymbol{S}_{\text{gt}}$. Although $\boldsymbol{S}_{\text{p}}$ and $\boldsymbol{S}_{\text{gt}}$ share the same surface geometry in the observed regions, their latents obtained by the VAE encoder in the corresponding regions show an obvious difference. }
    \label{fig:motivation}
\end{figure}

%% file: main.bbl
\begin{thebibliography}{89}
\providecommand{\natexlab}[1]{#1}
\providecommand{\url}[1]{\texttt{#1}}
\expandafter\ifx\csname urlstyle\endcsname\relax
  \providecommand{\doi}[1]{doi: #1}\else
  \providecommand{\doi}{doi: \begingroup \urlstyle{rm}\Url}\fi

\bibitem[Achlioptas et~al.(2017)Achlioptas, Diamanti, Mitliagkas, and Guibas]{mmd}
Panos Achlioptas, Olga Diamanti, Ioannis Mitliagkas, and Leonidas~J. Guibas.
\newblock Learning {R}epresentations and {G}enerative {M}odels for 3{D} {P}oint {C}louds.
\newblock In \emph{Proceedings of International Conference on Machine Learning (ICML)}, 2017.

\bibitem[Arun et~al.(1987)Arun, Huang, and Blostein]{icp}
K.~S. Arun, T.~S. Huang, and S.~D. Blostein.
\newblock Least-{S}quares {F}itting of {T}wo 3-{D} {P}oint {S}ets.
\newblock \emph{IEEE Transactions Pattern Analysis \& Machine Intelligence}, PAMI-9\penalty0 (5):\penalty0 698--700, 1987.

\bibitem[Cai et~al.(2022)Cai, Lin, Zhang, Wang, Wang, and Li]{cai2022learning}
Yingjie Cai, Kwan-Yee Lin, Chao Zhang, Qiang Wang, Xiaogang Wang, and Hongsheng Li.
\newblock Learning a {S}tructured {L}atent {S}pace for {U}nsupervised {P}oint {C}loud {C}ompletion.
\newblock In \emph{IEEE Conference on Computer Vision and Pattern Recognition (CVPR)}, pages 5543--5553, 2022.

\bibitem[{\c{C}}alli et~al.(2015){\c{C}}alli, Walsman, Singh, Srinivasa, Abbeel, and Dollar]{ycb}
Berk {\c{C}}alli, Aaron Walsman, Arjun Singh, Siddhartha~S. Srinivasa, Pieter Abbeel, and Aaron~M. Dollar.
\newblock Benchmarking in {M}anipulation {R}esearch: {T}he {YCB} {O}bject and {M}odel {S}et and {B}enchmarking {P}rotocols.
\newblock In \emph{International Conference on Advanced Robotics (ICAR)}, 2015.

\bibitem[Cao et~al.(2024)Cao, Lin, Wang, Huang, and Tan]{cao2024ssnerf}
Xiao Cao, Beibei Lin, Bo Wang, Zhiyong Huang, and Robby~T Tan.
\newblock Ssnerf: {S}parse {V}iew {S}emi-supervised {N}eural {R}adiance {F}ields with {A}ugmentation.
\newblock \emph{arXiv preprint arXiv:2408.09144}, 2024.

\bibitem[Cao et~al.(2025)Cao, Lin, Wang, Huang, and Tan]{cao20253dot}
Xiao Cao, Beibei Lin, Bo Wang, Zhiyong Huang, and Robby~T Tan.
\newblock 3{DOT}: {T}exture {T}ransfer for 3{DGS} {O}bjects from a {S}ingle {R}eference {I}mage.
\newblock \emph{arXiv preprint arXiv:2503.18853}, 2025.

\bibitem[Chang et~al.(2015)Chang, Funkhouser, Guibas, Hanrahan, Huang, Li, Savarese, Savva, Song, Su, et~al.]{chang2015shapenet}
Angel~X. Chang, Thomas Funkhouser, Leonidas~J. Guibas, Pat Hanrahan, Qixing Huang, Zimo Li, Silvio Savarese, Manolis Savva, Shuran Song, Hao Su, et~al.
\newblock {ShapeNet}: An {I}nformation-{R}ich {3D} {M}odel {R}epository.
\newblock \emph{arXiv preprint arXiv:1512.03012}, 2015.

\bibitem[Chen et~al.(2020)Chen, Chen, and Mitra]{chen2019unpaired}
Xuelin Chen, Baoquan Chen, and Niloy~J Mitra.
\newblock Unpaired {P}oint {C}loud {C}ompletion on {R}eal {S}cans {u}sing {A}dversarial {T}raining.
\newblock In \emph{International Conference on Learning Representations (ICLR)}, 2020.

\bibitem[Chen et~al.(2022)Chen, Li, Xie, and Wang]{chen2022single}
Ying-Hao Chen, Jian Li, Shi-Peng Xie, and Qin Wang.
\newblock Single-photon 3{D} {I}maging with a {M}ulti-stage {N}etwork.
\newblock \emph{Optics Express}, 30\penalty0 (16):\penalty0 29173--29188, 2022.

\bibitem[Choi et~al.(2016)Choi, Zhou, Miller, and Koltun]{redwood}
Sungjoon Choi, Qian-Yi Zhou, Stephen Miller, and Vladlen Koltun.
\newblock A {L}arge {D}ataset of {O}bject {S}cans.
\newblock \emph{arXiv:1602.02481}, 2016.

\bibitem[Chu et~al.(2025)Chu, Li, Wang, Ning, Lu, and Fan]{chu2025digging}
Jisheng Chu, Wenrui Li, Xingtao Wang, Kanglin Ning, Yidan Lu, and Xiaopeng Fan.
\newblock Digging into {I}ntrinsic {C}ontextual {I}nformation for {H}igh-fidelity 3{D} {P}oint {C}loud {C}ompletion.
\newblock In \emph{AAAI Conference on Artificial Intelligence (AAAI)}, pages 2573--2581, 2025.

\bibitem[Chu et~al.(2021)Chu, Pan, and Wang]{chu2021unsupervised}
Lei Chu, Hao Pan, and Wenping Wang.
\newblock Unsupervised {S}hape {C}ompletion via {D}eep {P}rior in the {N}eural {T}angent {K}ernel {P}erspective.
\newblock \emph{ACM Transactions on Graphics}, 40\penalty0 (3):\penalty0 1--17, 2021.

\bibitem[Chu et~al.(2023)Chu, Xie, Mo, Li, Nie{\ss}ner, Fu, and Jia]{chu2023diffcomplete}
Ruihang Chu, Enze Xie, Shentong Mo, Zhenguo Li, Matthias Nie{\ss}ner, Chi-Wing Fu, and Jiaya Jia.
\newblock Diff{C}omplete: {D}iffusion-based {G}enerative 3{D} {S}hape {C}ompletion.
\newblock In \emph{Conference on Neural Information Processing Systems (NeurIPS)}, pages 75951--75966, 2023.

\bibitem[Cui et~al.(2023)Cui, Qiu, Anwar, Liu, Xing, Zhang, and Barnes]{cui2023p2c}
Ruikai Cui, Shi Qiu, Saeed Anwar, Jiawei Liu, Chaoyue Xing, Jing Zhang, and Nick Barnes.
\newblock P2{C}: Self-supervised {P}oint {C}loud {C}ompletion from {S}ingle {P}artial {C}louds.
\newblock In \emph{IEEE International Conference on Computer Vision (ICCV)}, pages 14351--14360, 2023.

\bibitem[Dai et~al.(2017{\natexlab{a}})Dai, Chang, Savva, Halber, Funkhouser, and Nießner]{dataset_scannet}
Angela Dai, Angel~X. Chang, Manolis Savva, Maciej Halber, Thomas Funkhouser, and Matthias Nießner.
\newblock Scan{N}et: {R}ichly-{A}nnotated 3{D } {R}econstructions of {I}ndoor {S}cenes.
\newblock In \emph{2017 IEEE Conference on Computer Vision and Pattern Recognition (CVPR)}, pages 2432--2443, 2017{\natexlab{a}}.

\bibitem[Dai et~al.(2017{\natexlab{b}})Dai, Ruizhongtai~Qi, and Nie{\ss}ner]{dai2017shape}
Angela Dai, Charles Ruizhongtai~Qi, and Matthias Nie{\ss}ner.
\newblock {S}hape {C}ompletion using 3{D}-{E}ncoder-{P}redictor {C}{N}{N}s and {S}hape {S}ynthesis.
\newblock In \emph{IEEE Conference on Computer Vision and Pattern Recognition (CVPR)}, pages 5868--5877, 2017{\natexlab{b}}.

\bibitem[Deitke et~al.(2023{\natexlab{a}})Deitke, Liu, Wallingford, Ngo, Michel, Kusupati, Fan, Laforte, Voleti, Gadre, et~al.]{deitke2023objaversexl}
Matt Deitke, Ruoshi Liu, Matthew Wallingford, Huong Ngo, Oscar Michel, Aditya Kusupati, Alan Fan, Christian Laforte, Vikram Voleti, Samir~Yitzhak Gadre, et~al.
\newblock Objaverse-{XL}: {A} {U}niverse of 10{M}+ 3{D} {O}bjects.
\newblock \emph{arXiv preprint arXiv:2307.05663}, 2023{\natexlab{a}}.

\bibitem[Deitke et~al.(2023{\natexlab{b}})Deitke, Schwenk, Salvador, Weihs, Michel, VanderBilt, Schmidt, Ehsani, Kembhavi, and Farhadi]{deitke2023objaverse}
Matt Deitke, Dustin Schwenk, Jordi Salvador, Luca Weihs, Oscar Michel, Eli VanderBilt, Ludwig Schmidt, Kiana Ehsani, Aniruddha Kembhavi, and Ali Farhadi.
\newblock Objaverse: {A} {U}niverse of {A}nnotated 3{D} {O}bjects.
\newblock In \emph{IEEE Conference on Computer Vision and Pattern Recognition (CVPR)}, pages 13142--13153, 2023{\natexlab{b}}.

\bibitem[Du et~al.(2025)Du, Hu, Li, Xu, Huang, Fu, and Liu]{du2025hierarchical}
Keyu Du, Jingyu Hu, Haipeng Li, Hao Xu, Haibin Huang, Chi-Wing Fu, and Shuaicheng Liu.
\newblock Hierarchical {N}eural {S}emantic {R}epresentation for 3{D} {S}emantic {C}orrespondence.
\newblock In \emph{Proceedings of SIGGRAPH Asia}, pages 1--11, 2025.

\bibitem[Evans(1996)]{Evans1996}
James~D. Evans.
\newblock \emph{Straightforward Statistics for the Behavioral Sciences}.
\newblock Brooks/Cole, Pacific Grove, CA, 1996.

\bibitem[Geiger(2012)]{dataset_kitti}
Andreas Geiger.
\newblock Are {W}e {R}eady for {A}utonomous {D}riving? {T}he {KITTI} {V}ision {B}enchmark {S}uite.
\newblock In \emph{Proceedings of the 2012 IEEE Conference on Computer Vision and Pattern Recognition (CVPR)}, page 3354–3361, 2012.

\bibitem[He et~al.(2025)He, Zou, Chen, Guo, Liang, Yuan, Ouyang, Cao, and Li]{he2025triposf}
Xianglong He, Zi-Xin Zou, Chia-Hao Chen, Yuan-Chen Guo, Ding Liang, Chun Yuan, Wanli Ouyang, Yan-Pei Cao, and Yangguang Li.
\newblock Sparse{F}lex: {H}igh-{R}esolution and {A}rbitrary-{T}opology 3{D} {S}hape {M}odeling.
\newblock In \emph{IEEE International Conference on Computer Vision (ICCV)}, 2025.

\bibitem[Ho and Salimans(2022)]{ho2022classifierfreediffusionguidance}
Jonathan Ho and Tim Salimans.
\newblock Classifier-free {D}iffusion {G}uidance.
\newblock \emph{arxiv preprint arxiv:2207.12598}, 2022.

\bibitem[Ho et~al.(2020)Ho, Jain, and Abbeel]{ho2020denoising}
Jonathan Ho, Ajay Jain, and Pieter Abbeel.
\newblock Denoising {D}iffusion {P}robabilistic {M}odels.
\newblock In \emph{Conference on Neural Information Processing Systems (NeurIPS)}, pages 6840--6851, 2020.

\bibitem[Hong et~al.(2023)Hong, Yavartanoo, Neshatavar, and Lee]{hong2023acl}
Sangmin Hong, Mohsen Yavartanoo, Reyhaneh Neshatavar, and Kyoung~Mu Lee.
\newblock {ACL}-{SPC}: {A}daptive {C}losed-{L}oop system for {S}elf-supervised {P}oint {C}loud {C}ompletion.
\newblock In \emph{IEEE Conference on Computer Vision and Pattern Recognition (CVPR)}, pages 9435--9444, 2023.

\bibitem[Hong et~al.(2024)Hong, Zhang, Gu, Bi, Zhou, Liu, Liu, Sunkavalli, Bui, and Tan]{hong2024lrmlargereconstructionmodel}
Yicong Hong, Kai Zhang, Jiuxiang Gu, Sai Bi, Yang Zhou, Difan Liu, Feng Liu, Kalyan Sunkavalli, Trung Bui, and Hao Tan.
\newblock {LRM}: {L}arge {R}econstruction {M}odel for {S}ingle {I}mage to 3{D}.
\newblock In \emph{International Conference on Learning Representations (ICLR)}, 2024.

\bibitem[Hu* et~al.(2023)Hu*, Hui*, Liu, Zhang, and Fu]{hu2023clipxplore}
Jingyu Hu*, Ka-Hei Hui*, Zhengzhe Liu, Hao Zhang, and Chi-Wing Fu.
\newblock {CLIPXP}lore: {C}oupled {CLIP} and {S}hape {S}paces for 3{D} {S}hape {E}xploration.
\newblock In \emph{Proceedings of SIGGRAPH Asia}, pages 1--12, 2023.

\bibitem[Hu et~al.(2024{\natexlab{a}})Hu, Hui, Liu, Li, and Fu]{hu2023neural}
Jingyu Hu, Ka-Hei Hui, Zhengzhe Liu, Ruihui Li, and Chi-Wing Fu.
\newblock Neural {W}avelet-domain {D}iffusion for 3{D} {S}hape {G}eneration, {I}nversion, and {M}anipulation.
\newblock \emph{ACM Transactions on Graphics (TOG)}, 42\penalty0 (6), 2024{\natexlab{a}}.

\bibitem[Hu et~al.(2024{\natexlab{b}})Hu, Hui, Liu, Zhang, and Fu]{hu2024_cnsedit}
Jingyu Hu, Ka-Hei Hui, Zhengzhe Liu, Hao Zhang, and Chi-Wing Fu.
\newblock {CNS}-{E}dit: 3{D} {S}hape {E}diting via {C}oupled {N}eural {S}hape {O}ptimization.
\newblock In \emph{Proceedings of SIGGRAPH}, pages 1--12, 2024{\natexlab{b}}.

\bibitem[Hu et~al.(2026)Hu, Hu, Hui, Li, Liu, Cohen-Or, and Fu]{hu2026pegasus3dpersonalizationgeometry}
Jingyu Hu, Bin Hu, Ka-Hei Hui, Haipeng Li, Zhengzhe Liu, Daniel Cohen-Or, and Chi-Wing Fu.
\newblock {PEGA}sus: 3{D} {P}ersonalization of {G}eometry and {A}ppearance.
\newblock \emph{arXiv preprint arXiv:2602.08198}, 2026.

\bibitem[Huang et~al.(2025)Huang, Yan, Zhao, and Lee]{compc}
Tianxin Huang, Zhiwen Yan, Yuyang Zhao, and Gim~H Lee.
\newblock Com{PC}: {C}ompleting a 3{D} {P}oint {C}loud with 2{D} {D}iffusion {P}riors.
\newblock In \emph{International Conference on Learning Representations (ICLR)}, pages 51765--51784, 2025.

\bibitem[Huang et~al.(2020)Huang, Yu, Xu, Ni, and Le]{huang2020pf}
Zitian Huang, Yikuan Yu, Jiawen Xu, Feng Ni, and Xinyi Le.
\newblock P{F}-{N}et: {P}oint {F}ractal {N}etwork for 3{D} {P}oint {C}loud {C}ompletion.
\newblock In \emph{IEEE Conference on Computer Vision and Pattern Recognition (CVPR)}, pages 7662--7670, 2020.

\bibitem[Hui et~al.(2022{\natexlab{a}})Hui, Li, Hu, and Fu]{hui2022neural}
Ka-Hei Hui, Ruihui Li, Jingyu Hu, and Chi-Wing Fu.
\newblock Neural {W}avelet-domain {D}iffusion for 3{D} {S}hape {G}eneration.
\newblock In \emph{Proceedings of SIGGRAPH Asia}, pages 1--9, 2022{\natexlab{a}}.

\bibitem[Hui et~al.(2022{\natexlab{b}})Hui, Li, Hu, and Fu]{hui2022template}
Ka-Hei Hui, Ruihui Li, Jingyu Hu, and Chi-Wing Fu.
\newblock Neural {T}emplate: {T}opology-aware {R}econstruction and {D}isentangled {G}eneration of 3{D} {M}eshes.
\newblock In \emph{IEEE Conference on Computer Vision and Pattern Recognition (CVPR)}, pages 18572--18582, 2022{\natexlab{b}}.

\bibitem[Iwase et~al.(2024)Iwase, Liu, Guizilini, Gaidon, Kitani, Ambruș, and Zakharov]{octmae}
Shun Iwase, Katherine Liu, Vitor Guizilini, Adrien Gaidon, Kris Kitani, Rareș Ambruș, and Sergey Zakharov.
\newblock Zero-{S}hot {M}ulti-{O}bject {S}cene {C}ompletion.
\newblock In \emph{ECCV}, 2024.

\bibitem[Iwase et~al.(2025)Iwase, Irshad, Liu, Guizilini, Lee, Ikeda, Ayako, Nishiwaki, Kitani, Ambru{\c{s}}, and Zakharov]{zero-grasp}
Shun Iwase, Muhammad~Zubair Irshad, Katherine Liu, Vitor Guizilini, Robert Lee, Takuya Ikeda, Amma Ayako, Koichi Nishiwaki, Kris Kitani, Rare{\c{s}} Ambru{\c{s}}, and Sergey Zakharov.
\newblock {ZeroGrasp: {Z}ero-{S}hot {S}hape {R}econstruction {E}nabled {R}obotic {G}rasping}.
\newblock In \emph{CVPR}, 2025.

\bibitem[Kasten et~al.(2023)Kasten, Rahamim, and Chechik]{sdscomplete}
Yoni Kasten, Ohad Rahamim, and Gal Chechik.
\newblock Point {C}loud {C}ompletion with {P}retrained {T}ext-to-image {D}iffusion {M}odels.
\newblock In \emph{Conference on Neural Information Processing Systems (NeurIPS)}, pages 12171--12191, 2023.

\bibitem[Kim et~al.(2023)Kim, Kwon, Yang, and Yoon]{kim2023learning}
Jihun Kim, Hyeokjun Kwon, Yunseo Yang, and Kuk-Jin Yoon.
\newblock Learning {P}oint {C}loud {C}ompletion without {C}omplete {P}oint {C}louds: {A} {P}ose-aware {A}pproach.
\newblock In \emph{IEEE International Conference on Computer Vision (ICCV)}, pages 14157--14167, 2023.

\bibitem[Kim et~al.(2025)Kim, Kim, and Ye]{Kim_2025_ICCV}
Jeongsol Kim, Bryan~Sangwoo Kim, and Jong~Chul Ye.
\newblock Flow{DPS} : {F}low-{D}riven {P}osterior {S}ampling for {I}nverse {P}roblems.
\newblock In \emph{Proceedings of the IEEE/CVF International Conference on Computer Vision (ICCV)}, pages 12328--12337, 2025.

\bibitem[Li et~al.(2025{\natexlab{a}})Li, Zhu, and Wei]{genpc}
An Li, Zhe Zhu, and Mingqiang Wei.
\newblock Gen{PC}: {Z}ero-shot {P}oint {C}loud {C}ompletion via 3{D} {G}enerative {P}riors.
\newblock In \emph{Proceedings of the IEEE/CVF Conference on Computer Vision and Pattern Recognition (CVPR)}, pages 1308--1318, 2025{\natexlab{a}}.

\bibitem[Li et~al.(2023)Li, Gao, Tan, and Wei]{li2023proxyformer}
Shanshan Li, Pan Gao, Xiaoyang Tan, and Mingqiang Wei.
\newblock Proxy{F}ormer: {P}roxy {A}lignment {A}ssisted {P}oint {C}loud {C}ompletion with {M}issing {P}art {S}ensitive {T}ransformer.
\newblock In \emph{IEEE Conference on Computer Vision and Pattern Recognition (CVPR)}, pages 9466--9475, 2023.

\bibitem[Li et~al.(2026)Li, Tan, and Tan]{bridgingdaynighttargetclass}
Shuwei Li, Lei Tan, and Robby~T. Tan.
\newblock Bridging {D}ay and {N}ight: {T}arget-{C}lass {H}allucination {S}uppression in {U}npaired {I}mage {T}ranslation, 2026.

\bibitem[Li et~al.(2025{\natexlab{b}})Li, Zou, Liu, Wang, Liang, Yu, Liu, Guo, Liang, Ouyang, et~al.]{li2025triposg}
Yangguang Li, Zi-Xin Zou, Zexiang Liu, Dehu Wang, Yuan Liang, Zhipeng Yu, Xingchao Liu, Yuan-Chen Guo, Ding Liang, Wanli Ouyang, et~al.
\newblock Tripo{SG}: {H}igh-{F}idelity 3{D} {S}hape {S}ynthesis using {L}arge-{S}cale {R}ectified {F}low {M}odels.
\newblock \emph{arXiv preprint arXiv:2502.06608}, 2025{\natexlab{b}}.

\bibitem[Lipman et~al.(2008)Lipman, Levin, and Cohen-Or]{lipman2008green}
Yaron Lipman, David Levin, and Daniel Cohen-Or.
\newblock Green {C}oordinates.
\newblock \emph{ACM Transactions on Graphics}, 27\penalty0 (3):\penalty0 1--10, 2008.

\bibitem[Liu et~al.(2024)Liu, Chhatkuli, Postels, Van~Gool, and Tombari]{liu2024self}
Mengya Liu, Ajad Chhatkuli, Janis Postels, Luc Van~Gool, and Federico Tombari.
\newblock {S}elf-supervised {S}hape {C}ompletion via {I}nvolution and {I}mplicit {C}orrespondences.
\newblock In \emph{European Conference on Computer Vision (ECCV)}, pages 212--229, 2024.

\bibitem[Liu et~al.(2023{\natexlab{a}})Liu, Wu, Van~Hoorick, Tokmakov, Zakharov, and Vondrick]{liu2023zero}
Ruoshi Liu, Rundi Wu, Basile Van~Hoorick, Pavel Tokmakov, Sergey Zakharov, and Carl Vondrick.
\newblock {Z}ero-1-to-3: {Z}ero-shot {O}ne {I}mage to 3{D} {O}bject.
\newblock In \emph{IEEE International Conference on Computer Vision (ICCV)}, pages 9298--9309, 2023{\natexlab{a}}.

\bibitem[Liu et~al.(2023{\natexlab{b}})Liu, Hu, Hui, Qi, Cohen-Or, and Fu]{liu2023exim}
Zhengzhe Liu, Jingyu Hu, Ka-Hei Hui, Xiaojuan Qi, Daniel Cohen-Or, and Chi-Wing Fu.
\newblock Exim: A hybrid explicit-implicit representation for text-guided 3d shape generation.
\newblock \emph{ACM Transactions on Graphics (SIGGRAPH Asia)}, 42\penalty0 (6):\penalty0 1--12, 2023{\natexlab{b}}.

\bibitem[Luo and Hu(2021)]{luo2021diffusion}
Shitong Luo and Wei Hu.
\newblock {D}iffusion {P}robabilistic {M}odels for {3D} {P}oint {C}loud {G}eneration.
\newblock In \emph{IEEE Conference on Computer Vision and Pattern Recognition (CVPR)}, pages 2837--2845, 2021.

\bibitem[Lyu et~al.(2022)Lyu, Kong, Xu, Pan, and Lin]{lyu2021conditional}
Zhaoyang Lyu, Zhifeng Kong, Xudong Xu, Liang Pan, and Dahua Lin.
\newblock {A} {C}onditional {P}oint {D}iffusion-{R}efinement {P}aradigm for 3{D} {P}oint {C}loud {C}ompletion.
\newblock In \emph{International Conference on Learning Representations (ICLR)}, 2022.

\bibitem[Minderer et~al.(2023)Minderer, Gritsenko, and Houlsby]{owlvit}
Matthias Minderer, Alexey Gritsenko, and Neil Houlsby.
\newblock Scaling {O}pen-{V}ocabulary {O}bject {D}etection.
\newblock In \emph{Conference on Neural Information Processing Systems (NeurIPS)}, 2023.

\bibitem[Morris et~al.(2020)Morris, Lifland, Lanchantin, Ji, and Qi]{morris-etal-2020-reevaluating}
John Morris, Eli Lifland, Jack Lanchantin, Yangfeng Ji, and Yanjun Qi.
\newblock Reevaluating {A}dversarial {E}xamples in {N}atural {L}anguage.
\newblock In \emph{Findings of the Association for Computational Linguistics: EMNLP 2020}, pages 3829--3839, 2020.

\bibitem[Pauly et~al.(2005)Pauly, Mitra, Giesen, Gross, and Guibas]{pauly2005example}
Mark Pauly, Niloy~J Mitra, Joachim Giesen, Markus~H Gross, and Leonidas~J Guibas.
\newblock {E}xample-{B}ased 3{D} {S}can {C}ompletion.
\newblock In \emph{Eurographics Symposium on Geometry Processing (SGP)}, 2005.

\bibitem[Pauly et~al.(2008)Pauly, Mitra, Wallner, Pottmann, and Guibas]{pauly2008discovering}
Mark Pauly, Niloy~J Mitra, Johannes Wallner, Helmut Pottmann, and Leonidas~J Guibas.
\newblock {D}iscovering {S}tructural {R}egularity in 3{D} {G}eometry.
\newblock In \emph{ACM Transactions on Graphics (SIGGRAPH)}, pages 1--11, 2008.

\bibitem[Peng et~al.(2021)Peng, Jiang, Liao, Niemeyer, Pollefeys, and Geiger]{Peng2021SAP}
Songyou Peng, Chiyu~"Max" Jiang, Yiyi Liao, Michael Niemeyer, Marc Pollefeys, and Andreas Geiger.
\newblock {S}hape {A}s {P}oints: {A} {D}ifferentiable {P}oisson {S}olver.
\newblock In \emph{Conference on Neural Information Processing Systems (NeurIPS)}, 2021.

\bibitem[Ravi et~al.(2025)Ravi, Gabeur, Hu, Hu, Ryali, Ma, Khedr, R{\"a}dle, Rolland, Gustafson, Mintun, Pan, Alwala, Carion, Wu, Girshick, Dollar, and Feichtenhofer]{sam2}
Nikhila Ravi, Valentin Gabeur, Yuan-Ting Hu, Ronghang Hu, Chaitanya Ryali, Tengyu Ma, Haitham Khedr, Roman R{\"a}dle, Chloe Rolland, Laura Gustafson, Eric Mintun, Junting Pan, Kalyan~Vasudev Alwala, Nicolas Carion, Chao-Yuan Wu, Ross Girshick, Piotr Dollar, and Christoph Feichtenhofer.
\newblock {SAM} 2: {S}egment {A}nything in {I}mages and {V}ideos.
\newblock In \emph{International Conference on Learning Representations (ICLR)}, 2025.

\bibitem[Rombach et~al.(2022)Rombach, Blattmann, Lorenz, Esser, and Ommer]{rombach2022high}
Robin Rombach, Andreas Blattmann, Dominik Lorenz, Patrick Esser, and Bj{\"o}rn Ommer.
\newblock {H}igh-{R}esolution {I}mage {S}ynthesis with {L}atent {D}iffusion {M}odels.
\newblock In \emph{IEEE Conference on Computer Vision and Pattern Recognition (CVPR)}, pages 10684--10695, 2022.

\bibitem[Rubner et~al.(2000)Rubner, Tomasi, and Guibas]{emd2000}
Yossi Rubner, Carlo Tomasi, and Leonidas~J. Guibas.
\newblock The {E}arth {M}over's {D}istance as a {M}etric for {I}mage {R}etrieval.
\newblock \emph{International Journal Computer Vision}, 2000.

\bibitem[Shen et~al.(2012)Shen, Fu, Chen, and Hu]{shen2012structure}
Chao-Hui Shen, Hongbo Fu, Kang Chen, and Shi-Min Hu.
\newblock {S}tructure {R}ecovery by {P}art {A}ssembly.
\newblock \emph{ACM Transactions on Graphics (SIGGRAPH)}, 31\penalty0 (6):\penalty0 1--11, 2012.

\bibitem[Sung et~al.(2015)Sung, Kim, Angst, and Guibas]{sung2015data}
Minhyuk Sung, Vladimir~G Kim, Roland Angst, and Leonidas Guibas.
\newblock {D}ata-driven {S}tructural {P}riors for {S}hape {C}ompletion.
\newblock \emph{ACM Transactions on Graphics (SIGGRAPH)}, 34\penalty0 (6):\penalty0 1--11, 2015.

\bibitem[Tang et~al.(2024)Tang, Chen, Chen, Wang, Zeng, and Liu]{tang2024lgm}
Jiaxiang Tang, Zhaoxi Chen, Xiaokang Chen, Tengfei Wang, Gang Zeng, and Ziwei Liu.
\newblock {LGM}: {L}arge {M}ulti-{V}iew {G}aussian {M}odel for {H}igh-{R}esolution 3{D} {C}ontent {C}reation.
\newblock In \emph{European Conference on Computer Vision (ECCV)}, pages 1--18, 2024.

\bibitem[Thrun and Wegbreit(2005)]{thrun2005shape}
Sebastian Thrun and Ben Wegbreit.
\newblock Shape from {S}ymmetry.
\newblock In \emph{IEEE International Conference on Computer Vision (ICCV)}, pages 1824--1831, 2005.

\bibitem[Wang et~al.(2024)Wang, Cui, Guo, Li, Liu, and Shen]{wang2024pointattn}
Jun Wang, Ying Cui, Dongyan Guo, Junxia Li, Qingshan Liu, and Chunhua Shen.
\newblock Point{A}tt{N}: {Y}ou only {N}eed {A}ttention for {P}oint {C}loud {C}ompletion.
\newblock In \emph{AAAI Conference on Artificial Intelligence (AAAI)}, pages 5472--5480, 2024.

\bibitem[Wang et~al.(2025)Wang, He, Lv, Zhou, Xu, Yu, and Gu]{partnextnext}
Penghao Wang, Yiyang He, Xin Lv, Yukai Zhou, Lan Xu, Jingyi Yu, and Jiayuan Gu.
\newblock Part{N}e{X}t: {A} {N}ext-{G}eneration {D}ataset for {F}ine-{G}rained and {H}ierarchical 3{D} {P}art {U}nderstanding.
\newblock In \emph{Conference on Neural Information Processing Systems (NeurIPS)}, 2025.

\bibitem[Wang et~al.(2020)Wang, Ang~Jr, and Lee]{wang2020cascaded}
Xiaogang Wang, Marcelo~H Ang~Jr, and Gim~Hee Lee.
\newblock {C}ascaded {R}efinement {N}etwork for {P}oint {C}loud {C}ompletion.
\newblock In \emph{IEEE Conference on Computer Vision and Pattern Recognition (CVPR)}, pages 790--799, 2020.

\bibitem[Wei et~al.(2025)Wei, Feng, Ma, Wang, Zhou, and Li]{pcdreamer}
Guangshun Wei, Yuan Feng, Long Ma, Chen Wang, Yuanfeng Zhou, and Changjian Li.
\newblock {P}{C}{D}reamer: {P}oint {C}loud {C}ompletion {T}hrough {M}ulti-view {D}iffusion {P}riors.
\newblock In \emph{Proceedings of the IEEE/CVF Conference on Computer Vision and Pattern Recognition (CVPR)}, pages 27243--27253, 2025.

\bibitem[Wen et~al.(2021{\natexlab{a}})Wen, Han, Cao, Wan, Zheng, and Liu]{wen2021cycle4completion}
Xin Wen, Zhizhong Han, Yan-Pei Cao, Pengfei Wan, Wen Zheng, and Yu-Shen Liu.
\newblock {C}ycle4{C}ompletion: {U}npaired {P}oint {C}loud {C}ompletion using {C}ycle {T}ransformation with {M}issing {R}egion {C}oding.
\newblock In \emph{IEEE Conference on Computer Vision and Pattern Recognition (CVPR)}, pages 13080--13089, 2021{\natexlab{a}}.

\bibitem[Wen et~al.(2021{\natexlab{b}})Wen, Xiang, Han, Cao, Wan, Zheng, and Liu]{wen2021pmp}
Xin Wen, Peng Xiang, Zhizhong Han, Yan-Pei Cao, Pengfei Wan, Wen Zheng, and Yu-Shen Liu.
\newblock {P}{M}{P}-{N}et: {P}oint {C}loud {C}ompletion by learning {M}ulti-step {P}oint {M}oving {P}aths.
\newblock In \emph{IEEE Conference on Computer Vision and Pattern Recognition (CVPR)}, pages 7443--7452, 2021{\natexlab{b}}.

\bibitem[Wu et~al.(2020)Wu, Chen, Zhuang, and Chen]{wu2020multimodal}
Rundi Wu, Xuelin Chen, Yixin Zhuang, and Baoquan Chen.
\newblock {M}ultimodal {S}hape {C}ompletion via {C}onditional {G}enerative {A}dversarial {N}etworks.
\newblock In \emph{European Conference on Computer Vision (ECCV)}, pages 281--296, 2020.

\bibitem[Wu et~al.(2025)Wu, Lin, Zhang, Zeng, Yang, Bao, Qian, Zhu, Torr, Cao, and Yao]{wu2025direct3ds2gigascale3dgeneration}
Shuang Wu, Youtian Lin, Feihu Zhang, Yifei Zeng, Yikang Yang, Yajie Bao, Jiachen Qian, Siyu Zhu, Philip Torr, Xun Cao, and Yao Yao.
\newblock {D}irect3{D}-{S}2: {G}igascale 3{D} {G}eneration {M}ade {E}asy with {S}patial {S}parse {A}ttention.
\newblock In \emph{Conference on Neural Information Processing Systems (NeurIPS)}, 2025.

\bibitem[Xiang et~al.(2025)Xiang, Lv, Xu, Deng, Wang, Zhang, Chen, Tong, and Yang]{xiang2025structured}
Jianfeng Xiang, Zelong Lv, Sicheng Xu, Yu Deng, Ruicheng Wang, Bowen Zhang, Dong Chen, Xin Tong, and Jiaolong Yang.
\newblock {S}tructured 3{D} {L}atents for {S}calable and {V}ersatile 3{D} {G}eneration.
\newblock In \emph{IEEE Conference on Computer Vision and Pattern Recognition (CVPR)}, pages 21469--21480, 2025.

\bibitem[Xiang et~al.(2021)Xiang, Wen, Liu, Cao, Wan, Zheng, and Han]{xiang2021snowflakenet}
Peng Xiang, Xin Wen, Yu-Shen Liu, Yan-Pei Cao, Pengfei Wan, Wen Zheng, and Zhizhong Han.
\newblock {{S}nowflake{N}et}: {P}oint {C}loud {C}ompletion by {S}nowflake {P}oint {D}econvolution with {S}kip-{T}ransformer.
\newblock In \emph{IEEE Conference on Computer Vision and Pattern Recognition (CVPR)}, pages 5499--5509, 2021.

\bibitem[Xie et~al.(2021)Xie, Wang, Zhang, Yang, Chen, and Wen]{xie2021style}
Chulin Xie, Chuxin Wang, Bo Zhang, Hao Yang, Dong Chen, and Fang Wen.
\newblock {S}tyle-based {P}oint {G}enerator with {A}dversarial {R}endering for {P}oint {C}loud {C}ompletion.
\newblock In \emph{IEEE Conference on Computer Vision and Pattern Recognition (CVPR)}, pages 4619--4628, 2021.

\bibitem[Xie et~al.(2020)Xie, Yao, Zhou, Mao, Zhang, and Sun]{xie2020grnet}
Haozhe Xie, Hongxun Yao, Shangchen Zhou, Jiageng Mao, Shengping Zhang, and Wenxiu Sun.
\newblock {G}{R}{N}et: {G}ridding {R}esidual {N}etwork for {D}ense {P}oint {C}loud {C}ompletion.
\newblock In \emph{European Conference on Computer Vision (ECCV)}, pages 365--381, 2020.

\bibitem[Yan et~al.(2023)Yan, Tan, Zeng, and Liu]{RS-HM}
Weilong Yan, Robby~T. Tan, Bing Zeng, and Shuaicheng Liu.
\newblock Deep {H}omography {M}ixture for {S}ingle {I}mage {R}olling {S}hutter {C}orrection.
\newblock In \emph{Proceedings of the IEEE/CVF International Conference on Computer Vision (ICCV)}, pages 9868--9877, 2023.

\bibitem[Yan et~al.(2025)Yan, Li, Li, Shao, and Tan]{syn2real-depth}
Weilong Yan, Ming Li, Haipeng Li, Shuwei Shao, and Robby~T. Tan.
\newblock Synthetic-to-{R}eal {S}elf-supervised {R}obust {D}epth {E}stimation via {L}earning with {M}otion and {S}tructure {P}riors.
\newblock In \emph{Proceedings of the IEEE/CVF Conference on Computer Vision and Pattern Recognition (CVPR)}, pages 21880--21890, 2025.

\bibitem[Yan et~al.(2022)Yan, Lin, Mitra, Lischinski, Cohen-Or, and Huang]{yan2022shapeformer}
Xingguang Yan, Liqiang Lin, Niloy~J. Mitra, Dani Lischinski, Daniel Cohen-Or, and Hui Huang.
\newblock {S}hape{F}ormer: {S}hapelet {T}ransformer for {M}ultivariate {T}ime {S}eries {C}lassification.
\newblock In \emph{IEEE Conference on Computer Vision and Pattern Recognition (CVPR)}, pages 6239--6249, 2022.

\bibitem[Yoo and Qi(2021)]{yoo-qi-2021-towards-improving}
Jin~Yong Yoo and Yanjun Qi.
\newblock Towards {I}mproving {A}dversarial {T}raining of {NLP} {M}odels.
\newblock In \emph{Findings of the Association for Computational Linguistics: EMNLP 2021}, pages 945--956, 2021.

\bibitem[Yu et~al.(2025)Yu, Duan, Herrmann, Freeman, and Wu]{wonderworld}
Hong-Xing Yu, Haoyi Duan, Charles Herrmann, William~T. Freeman, and Jiajun Wu.
\newblock Wonder{W}orld: {I}nteractive 3{D} {S}cene {G}eneration {f}rom a {S}ingle {I}mage.
\newblock In \emph{Proceedings of the IEEE/CVF Conference on Computer Vision and Pattern Recognition (CVPR)}, pages 5916--5926, 2025.

\bibitem[Yu et~al.(2021)Yu, Rao, Wang, Liu, Lu, and Zhou]{yu2021pointr}
Xumin Yu, Yongming Rao, Ziyi Wang, Zuyan Liu, Jiwen Lu, and Jie Zhou.
\newblock {P}oin{T}r: {D}iverse {P}oint {C}loud {C}ompletion with {G}eometry-{A}ware {T}ransformers.
\newblock In \emph{IEEE Conference on Computer Vision and Pattern Recognition (CVPR)}, pages 12498--12507, 2021.

\bibitem[Yu et~al.(2023)Yu, Rao, Wang, Lu, and Zhou]{adapointr}
Xumin Yu, Yongming Rao, Ziyi Wang, Jiwen Lu, and Jie Zhou.
\newblock {A}da{P}oin{T}r: {D}iverse {P}oint {C}loud {C}ompletion with {A}daptive {G}eometry-{A}ware {T}ransformers.
\newblock \emph{IEEE Transactions on Pattern Analysis and Machine Intelligence}, 45\penalty0 (12):\penalty0 14114--14130, 2023.

\bibitem[Yuan et~al.(2018)Yuan, Khot, Held, Mertz, and Hebert]{yuan2018pcn}
Wentao Yuan, Tejas Khot, David Held, Christoph Mertz, and Martial Hebert.
\newblock {P}{C}{N}: {P}oint {C}ompletion {N}etwork.
\newblock In \emph{International Conference on 3D Vision (3DV)}, pages 728--737, 2018.

\bibitem[Zhang et~al.(2021)Zhang, Chen, Cai, Pan, Zhao, Yi, Yeo, Dai, and Loy]{shapeinv}
Junzhe Zhang, Xinyi Chen, Zhongang Cai, Liang Pan, Haiyu Zhao, Shuai Yi, Chai~Kiat Yeo, Bo Dai, and Chen~Change Loy.
\newblock {U}nsupervised 3{D} {S}hape {C}ompletion through {G}{A}{N} {I}nversion.
\newblock In \emph{IEEE Conference on Computer Vision and Pattern Recognition (CVPR)}, 2021.

\bibitem[Zhang et~al.(2024)Zhang, Wang, Zhang, Qiu, Pang, Jiang, Yang, Xu, and Yu]{zhang2024clay}
Longwen Zhang, Ziyu Wang, Qixuan Zhang, Qiwei Qiu, Anqi Pang, Haoran Jiang, Wei Yang, Lan Xu, and Jingyi Yu.
\newblock {C}{L}{A}{Y}: {A} {C}ontrollable {L}arge-scale {G}enerative {M}odel for {C}reating {H}igh-quality 3{D} {A}ssets.
\newblock \emph{ACM Transactions on Graphics (SIGGRAPH)}, 43\penalty0 (4):\penalty0 1--20, 2024.

\bibitem[Zhang et~al.(2025)Zhang, Feng, Chen, and Fan]{zhang2024dcpi}
Mengtan Zhang, Yi Feng, Qijun Chen, and Rui Fan.
\newblock {DCPI-{D}epth}: {E}xplicitly {I}nfusing {D}ense {C}orrespondence {P}rior to {U}nsupervised {M}onocular {D}epth {D}stimation.
\newblock \emph{IEEE Transactions on Image Processing}, 34:\penalty0 4258--4272, 2025.

\bibitem[Zhang et~al.(2020)Zhang, Yan, and Xiao]{zhang2020detail}
Wenxiao Zhang, Qingan Yan, and Chunxia Xiao.
\newblock {D}etail {P}reserved {P}oint {C}loud {C}ompletion via {S}eparated {F}eature {A}ggregation.
\newblock In \emph{European Conference on Computer Vision (ECCV)}, pages 512--528, 2020.

\bibitem[Zhang et~al.(2026)Zhang, Yan, Shi, Qiu, He, Li, Li, and Fan]{4dpcchat}
Xindan Zhang, Weilong Yan, Yufei Shi, Xuerui Qiu, Tao He, Ying Li, Ming Li, and Hehe Fan.
\newblock 4{DPC}$^2$hat: {T}owards {D}ynamic {P}oint {C}loud {U}nderstanding with {F}ailure-{A}ware {B}ootstrapping.
\newblock \emph{arXiv preprint arXiv:2602.03890}, 2026.

\bibitem[Zhao et~al.(2025)Zhao, Lai, Lin, Zhao, Liu, Yang, Feng, Yang, Zhang, Yang, et~al.]{zhao2025hunyuan3d}
Zibo Zhao, Zeqiang Lai, Qingxiang Lin, Yunfei Zhao, Haolin Liu, Shuhui Yang, Yifei Feng, Mingxin Yang, Sheng Zhang, Xianghui Yang, et~al.
\newblock {H}unyuan3{D} 2.0: {S}caling {D}iffusion {M}odels for {H}igh {R}esolution {T}extured 3{D} {A}ssets {G}eneration.
\newblock \emph{arXiv preprint arXiv:2501.12202}, 2025.

\bibitem[Zhou et~al.(2021)Zhou, Du, and Wu]{zhou20213d}
Linqi Zhou, Yilun Du, and Jiajun Wu.
\newblock 3{D} {S}hape {G}eneration and {C}ompletion through {P}oint-{V}oxel {D}iffusion.
\newblock In \emph{IEEE International Conference on Computer Vision (ICCV)}, pages 5826--5835, 2021.

\bibitem[Zhu et~al.(2023)Zhu, Chen, He, Wang, Qin, and Wei]{zhu2023svdformer}
Zhe Zhu, Honghua Chen, Xing He, Weiming Wang, Jing Qin, and Mingqiang Wei.
\newblock {S}{V}{D}{F}ormer: {C}omplementing {P}oint {C}loud via {S}elf-view {A}ugmentation and {S}elf-structure {D}ual-{G}enerator.
\newblock In \emph{IEEE International Conference on Computer Vision (ICCV)}, pages 14508--14518, 2023.

\end{thebibliography}
